%% file: Arxiv-VIE-Full.tex
\def\year{2021}\relax
\documentclass[letterpaper]{article} 
\usepackage{aaai21}  
\usepackage{times}  
\usepackage{helvet} 
\usepackage{courier}  
\usepackage[hyphens]{url}  
\usepackage{graphicx} 
\usepackage{graphicx}  
\usepackage{natbib}  
\usepackage{caption} 
\nocopyright

\usepackage{siunitx}

\usepackage{amsmath}
\usepackage{amsthm,bm}
\usepackage{amssymb}
\usepackage{amsfonts}
\usepackage{multirow}
\usepackage{amssymb}
\usepackage[ruled]{algorithm2e}
\usepackage{subcaption}
\usepackage{multibib}

\newcites{SM}{SM References}
\usepackage{mwe}
\usepackage{booktabs,tabularx}
\usepackage[dvipsnames]{xcolor}

\usepackage{bbm}

\newcommand{\beginsupplement}{%
        \setcounter{table}{0}
        \renewcommand{\thetable}{S\arabic{table}}%
        \setcounter{figure}{0}
        \renewcommand{\thefigure}{S\arabic{figure}}%
     }

\newcommand{\EE}{\mathbb{E}}
\newcommand{\BR}{\mathbb{R}}
\newcommand{\ud}{\text{d}}
\newcommand{\KL}{\text{KL}}

\newcommand{\ELBO}{\text{ELBO}}

\newcommand{\beq}{\begin{equation}}
\newcommand{\eeq}{\end{equation}}
\newcommand{\beqs}{\begin{eqnarray}}
\newcommand{\eeqs}{\end{eqnarray}}
\newcommand{\barr}{\begin{array}}
\newcommand{\earr}{\end{array}}
\newcommand{\bal}{\begin{align}}
\newcommand{\eal}{\end{align}}

\begin{document}

\title{Variational Disentanglement for Rare Event Modeling}
\author {
        Zidi Xiu,
        Chenyang Tao,
        Michael Gao,
        Connor Davis,
        Benjamin Goldstein,
        Ricardo Henao
        \\
}
\affiliations {
    Duke University \\
    \{zidi.xiu, chenyang.tao\}@duke.edu
}








\maketitle

\begin{abstract}
Combining the increasing availability and abundance of healthcare data and the current advances in machine learning methods have created renewed opportunities to improve clinical decision support systems.
However, in healthcare risk prediction applications, the proportion of cases with the condition (label) of interest is often very low relative to the available sample size.
Though very prevalent in healthcare, such imbalanced classification settings are also common and challenging in many other scenarios.
So motivated, we propose a variational disentanglement approach to semi-parametrically learn from rare events in heavily imbalanced classification problems.
Specifically, we leverage the imposed extreme-distribution behavior on a latent space to extract information from low-prevalence events, 
and develop a robust prediction arm that joins the merits of the generalized additive model and isotonic neural nets. 
Results on synthetic studies and diverse real-world datasets, including mortality prediction on a COVID-19 cohort, demonstrate that the proposed approach outperforms existing alternatives.
\end{abstract}

\section{INTRODUCTION}
Early identification of in-hospital patients who are at imminent risk of life-threatening events, {\it e.g.}, death, ventilation or intensive care unit (ICU) transfer, is a critical subject in clinical care \shortcite{bedoya2019minimal}.
Especially during a pandemic like COVID-19, the needs for healthcare change dramatically.
With the ability to accurately predict the risk, an automated triage system will be well-positioned to help clinicians better allocate resources and attention to those patients whose adverse outcomes can be averted if early intervention efforts were in place.

Despite the great promise it holds, with the richness of modern Electronic Health Record (EHR) repositories, the construction of such a system faces practical challenges.
A major obstacle is the scarcity of patients experiencing adverse outcomes of interest.
In the COVID-19 scenario, which we consider in our experiments, the mortality of patients tested positive at the Duke University Health System (DUHS) is slightly lower than $3\%$.
Further, in another typical EHR dataset we consider, less than $5\%$ of patients are reported to suffer adverse outcomes (ICU transfer or death).
In these {\it low-prevalence} scenarios, commonly seen in clinical practice, standard classification models such as logistic regression suffer from {\it majority domination}, in which models tend to favor the prediction accuracy of majority groups.
This is clearly undesirable for critical-care applications, given the high false negative rates (Type-II error), in which patients in urgent need of care could be falsely categorized.

Situations where the distribution of labels is highly skewed and the accuracy of the minority class bears particular significance \shortcite{dal2017credit, lu2020evaluating, machado2020rare} have been associated with the name {\it imbalanced dataset} \shortcite{he2009learning}, whereas the methods dealing with such cases are coined {\it extreme classification} \shortcite{zong2013weighted}.
Under such a setting, the lack of representation of minority cases severely undermines the ability of a standard learner to discriminate, relative to balanced datasets
\shortcite{mitchell1999machine}.
Consequently, these solutions do not generalize well on minority classes, where the primary interest is usually focused.

To address such a dilemma, several remedies have been proposed to account for the imbalance between class representations.
One of the most popular strategies is the sampling-based adjustment, where during training, a model oversamples the minority classes (or undersamples the majority classes) to create balance artificially \shortcite{drummond2003c4}.
To overcome the biases and the lack of information that naive sampling adjustments might induce, variants have been proposed to maximally preserve the clustering structure of the original dataset \shortcite{mani2003knn, yen2009cluster} and to promote diversity of oversampling schemes \shortcite{han2005borderline}.
Alternatively, cost-sensitive weighting where minority losses are assigned larger weights
provides another popular option
via tuning the relative importance of minority classes \shortcite{elkan2001foundations,munro1996neural,zhou2005training}.
%


While the above two strategies introduce heuristics to alleviate the issues caused by class imbalance, importance sampling (IS) offers a principled treatment that flexibly combines the merits of the two \shortcite{hahn1987developments, heidelberger1995fast}.
Each example is sampled with the probability of a pre-specified importance weight, and with the weight's inverse
when accounting for the relative contribution in the overall loss.
This helps to flexibly tune the representation of rare events during  training, without biasing the data distribution \shortcite{heidelberger1995fast, shimodaira2000improving, gretton2009covariate}.
It is important to note that, poor choice of importance weights may result in uncontrolled variance  that destabilizes training \shortcite{robert2013monte, botev2008efficient}, calling for adaptive \shortcite{rubinstein2013cross} or variance reduction schemes \shortcite{rubinstein2016simulation} to protect against such degeneracy.  


Apart from the above strategies that fall within the standard empirical risk minimization framework, recent developments explicitly seek better generalization for the minority classes.
One such example is the {\it one-class classification} that aims to capture one target class from a general population  \shortcite{tax2002one}.
{\it Meta-learning} and {\it few-shot learning} strategies instead trying to transfer the knowledge learned from data-rich classes to facilitate the learning of data-scarce classes \shortcite{bohning2015meta,  finn2017model}.
Additionally, non-cross-entropy based losses or penalties have been proved useful to imbalanced classification tasks \shortcite{weinberger2009distance, huang2016learning}.
For instance, the Focal loss \shortcite{lin2017focal} up-weights the harder examples, and \citet{cao2019learning} introduced a label-distribution-aware margin loss encouraging minority classes having larger margins. 

In this work, we present a novel solution called {\it variational inference for extremals} (VIE), capitalizing on the learning of more generalizable representations for the minority classes.
Our proposal is motivated by the observation that the statistical features of ``rarity'' have been largely overlooked in the current literature of rare-event modeling. And the uncertainties of rare-events are often not considered.
Framed under the Variational Inference framework, we formulate our model with the assumption that the extreme presentation of (unobserved) latent variables can lead to the occurrence (or the inhibition) of rare events.
This encourages the accurate characterization of the tail distribution of the data representation, which has been missed by prior work to the best of our knowledge.
Building upon the state-of-the-art machine learning techniques, our solution features the following contributions: ($i$) the model accounts for representation uncertainty based on variational inference; ($ii$) the adoption of mixed Generalized Pareto priors to promote the learning of heavy-tailed feature representations; and ($iii$) integration of additive isotonic regression to disentangle representation and facilitate generalization.
We demonstrate how our framework facilitates both model generalization and interpretation, with strong empirical performance reported across a wide-range of benchmarks. 

\section{BACKGROUND}
To simplify our presentation, we focus on the problem of rare event classification for binary outcomes.
The generalization to the multiple-class scenario is simple and presented in the Supplementary Material (SM)
\footnote{SM can be found at \url{https://arxiv.org/abs/2009.08541}}.
Let $D=\{x_i, y_i\}_{i=1}^N$ be a dataset of interest, where $x_i$ and $y_i$ denote predictors and outcomes, respectively, and $N$ is the sample size.
Without loss of generality, we denote $y=1$ as the minority event label (indicating the occurrence of an event of interest), and $y=0$ as the majority label.

In the following, we will briefly review the three main techniques we used in this work, namely, {\it variational inference} (VI), {\it extreme value theory} (EVT), and {\it additive isotonic regression}.
VI allows for approximate maximum likelihood inference while accounting for data uncertainty.
EVT provides a principled and efficient way to model extreme, heavy-tailed representations.
Additive isotonic regression further introduces monotonic constraints to {\em disentangle} the contribution of each latent dimension to the outcome.
%

\subsection{Variational inference}\label{sec:vaeintro}
%
Consider a latent variable model $p_{\theta}(v,z)=p_{\theta}(v|z)p(z)$, where $v\in \BR^m$ is the observable data, $z\in \BR^p$ is the unobservable latent variable, and $\theta$ represents the parameters of the likelihood model, $p_{\theta}(v|z)$.
The marginal likelihood $p_{\theta}(v) = \int p_{\theta}(v,z) \ud z$ requires integrating out the latent $z$, which typically, for complex distributions, does not enjoy a closed-form expression.
This intractability prevents direct maximum likelihood estimation for $\theta$ in the latent variable setup.
To overcome this difficulty, Variational Inference (VI) optimizes computationally tractable variational bounds to the marginal log-likelihood \shortcite{kingma2013auto, chen2018variational}.
Concretely, the most popular choice of VI optimizes the following Evidence Lower Bound (ELBO):
%
\begin{align}
\text{ELBO}(v;p_{\theta}(v,z),q_{\phi}(z|v)) & \triangleq \EE_{Z\sim q_{\phi}(z|v)}
\left[ \log \frac{p_{\theta}(v,Z)}{q_{\phi}(Z|v)} \right] \notag \\
& \leq \log p_{\theta}(v) , \label{eq:elbo}
\end{align}
%
where $q_\phi(z|v)$ is an approximation to the true (unknown) posterior $p_{\theta}(z|v)$, and the inequality is a direct result of Jensen's inequality.
The variational gap between the ELBO and true marginal $\log$-likelihood, {\it i.e.}, $\log p_{\theta}(v) - \text{ELBO}(v;p_{\theta}(v,z),q_{\phi}(z|v))$, is given by the 
Kullback–Leibler (KL) divergence between posteriors, 
{\it i.e.}, $\KL(q_{\phi}(z|v)||p_{\theta}(z|v)) = \EE_{ Z\sim q_{\phi}(z|v)}[\log q_{\phi}(Z|v)] - \EE_{ Z\sim q_{\phi}(z|v)}[\log p_{\theta}(Z|v)]$,
which implies that the ELBO tightens as $q_{\phi}(z|v)$ approaching the true posterior $p_{\theta}(z|v)$.
For estimation, we seek parameters $\theta$ and $\phi$ that maximize the ELBO in \eqref{eq:elbo}.

Given a set of observations $\{ v_i \}_{i=1}^N$ sampled from data distribution $v\sim p_{d}(v)$, maximizing the expected ELBO is also equivalent to minimizing the KL divergence $\KL(p_d(v) \parallel p_{\theta}(v))$ between the empirical and model distributions.
When $p_{\theta}(v|z)$ and $q_{\phi}(z|v)$ are specified as neural networks, the resulting architecture is commonly known as the {\it variational auto-encoder} (VAE) \shortcite{kingma2013auto}, where $q_{\phi}(z|v)$ and $p_{\theta}(v|z)$ and are known as {\em encoder} and {\em decoder}, respectively. 
Note that $q_{\phi}(z|v)$ is often used for subsequent inference tasks on new data.

\subsection{Extreme Value Theory}
%
Extreme Value Theory (EVT) provides a principled probabilistic framework for describing events with extremely low probabilities, which we seek to exploit for better rare event modeling.
In particular, we focus on the {\it exceedance} models, where we aim to capture the asymptotic statistical behavior of values surpassing an extreme threshold \shortcite{davison1990models, tao2017generalized}, 
which we briefly review below following the notation of \citet{coles2001introduction}.
Without loss of generality, we consider exceedance to the right, {\it i.e.}, values greater than a threshold $u$. 
For a random variable $X$, the conditional cumulative distribution of exceedance level $x$ beyond $u$ is given by $F_u(x) = P(X-u\le x|X>u) = \frac{F(x+u)-F(u)}{1-F(u)}$,
where $x>0$ and $F(x)$ denotes the cumulative density function for $X$.

A major result from EVT is that under some mild regularity conditions, {\it e.g.}, continuity at the right end of $F(x)$ and others, $F_u(x)$ will converge to the family of Generalized Pareto Distributions (GPD) regardless of $F(x)$, as $u$ approaches the right support boundary of $F(x)$ \shortcite{balkema1974residual, pickands1975statistical},
{\it i.e.}, $\lim_{u\rightarrow\infty} F_u(x) \stackrel{L_{\infty}}{\longrightarrow} G_{\xi,\sigma, u}(x)$ \shortcite{falk2010laws}, where 
$\text{GPD}_{\xi ,\sigma, u}(x)$ is of the form
\begin{align}
    G_{\xi ,\sigma, u}(x)= 
\begin{cases}
    1-[1+\xi (x-u)/\sigma]^{-\frac{1}{\xi}},& \text{if }\xi \ne 0 \\
    1-\exp[-(x-u)/\sigma],              & \text{if }\xi = 0
\end{cases}
\end{align}
where $\sigma$ is a positive scale parameter.
When $\xi < 0$ the exceedance $x$ has bounded support $0\le x \le u-\sigma/\xi$, otherwise when $\xi\ge 0$, $x$ is unbounded. 
A major implication of this asymptotic behavior is that, 
for modeling extreme values, one only needs to fit extreme samples to the log-likelihood function of the GPD. 

\subsection{Additive Isotonic Regression}
Also known as monotonic regression, isotonic regression is a non-parametric regression model that constrains the relation between predictor and outcome to be monotonic, ({\it e.g.}, non-decreasing $f(a) \leq f(b)$ for $a\leq b$) \shortcite{barlow1972statistical}.
Such monotonic constraint is a natural and flexible extension to the standard linear relation assumed by many statistical models.
To accommodate multi-covariate predictors, additive isotonic regression combines isotonic models for each individual one-dimensional predictor 
\shortcite{bacchetti1989additive}.
Standard implementations often involve specialized algorithms, such as 
local scoring algorithms \shortcite{hastie2017generalized} and the alternating conditional expectation (ACE) method of \citet{breiman1985estimating}.
All these approaches typically require costly iterative computations and are not scalable to large datasets.
Here we consider recent advances in unconstrained monotonic neural networks, which allow for efficient and flexible end-to-end learning of monotonic relations with robust neural nets based on standard training schemes such as stochastic gradient descent \shortcite{sill1998monotonic, wehenkel2019unconstrained}.

\section{VARIATIONAL INFERENCE OF EXTREMALS}
%
%
%
The proposed model is based on the hypothesis that 
\textit{extreme events are driven by the extreme values of some latent factors}.
Specifically, we propose to recast the learning of low-prevalence events into the learning of extreme latent representations, thus amortizing the difficulties associated with directly modeling rare events as outcomes.
To allow for more efficient learning from the rare events, we make some further assumptions to regularize the latent representation: ($i$) {\it effect disentanglement}: the contribution from each dimension of the latent representation to the event occurrence is additive; ($ii$) {\it effect monotonicity}: there is a monotonic relation between the outcome likelihood and the values of each dimension of the latent representation.
The key to the proposed approach is using an additive isotonic neural network to model the one-dimensional disentangled monotonic relations from a latent representation, which is obtained via variational inference.
Specifically, we impose an EVT prior to explicitly capture the information from the few minority group samples into the tail behavior of the extreme representation.
Below we provide the rationale for our choices followed by a description of all model components. 

{\bf Disentanglement \& additive isotonic regression.}
Consistent with assumptions ($i$) and ($ii$), we posit a scenario in which the underlying representation of extreme events is more frequent at the far end of the representation spectrum, for which additive isotonic regression is ideal.
The disentanglement consists of modeling each latent dimension individually, thus avoiding the curse of dimensionality when modeling combinatorial effects with few examples.
Further, the monotonicity constraint imposed by the isotonic regression model restricts possible effect relations, thereby improving generalization error by learning with a smaller, yet still sufficiently expressive, class of models \shortcite{bacchetti1989additive}.

{\bf EVT \& VI.}
Note that the spread of representation of extreme events is expected to be more uncertain relative to those of the normal, more abundant events, due to a few plausible causes: ($i$) extreme events represent the breakdown of system normality and are expected to behave in uncertain ways; ($ii$) there is only a small number of examples available for the extreme events, so the learned feature encoder will tend to be unreliable.
As a result, it is safely expected that the encoded features associated with the extremes events will lie outside the effective support of the Gaussian distribution assumed by the standard VI model. 
In other words, the representation of the events can manifest as a heavy-tailed distribution.
This will compromise the validity and generalizability of a prediction model if not dealt with appropriately.
So motivated, we explicitly model the distribution of the extreme underlying representations via EVT. 
Using EVT, we decouple the learning of the tail end of the representation distribution.
Since EVT-based estimation only requires very few parameters, it allows for accurate modeling with a small set of tail-end samples.
Further, in combination with the variational inference framework, it accounts for representation uncertainty via the use of a stochastic encoder, which further strengthens model robustness. 

\begin{figure}[t!]
    \centering
    \includegraphics[width=\linewidth]{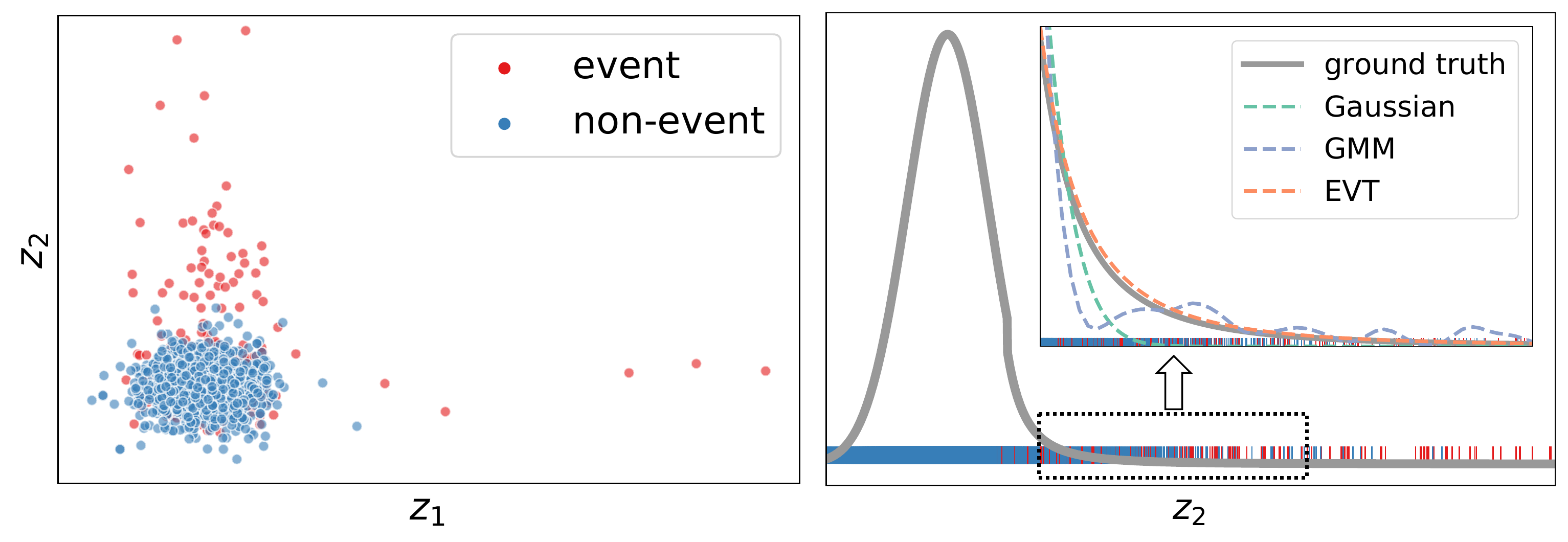}
    \caption{Left: Distribution of a two-dimensional latent space $z$ where the long tail associates with higher risk. Right: Tail estimations with different schemes for the long-tailed data in one-dimensional space. EVT provides more accurate characterization comparing to other mechanisms.}
    \label{fig:VItailestimation}
\end{figure}

{\bf Benefits of heavy-tailed modeling.}
A few other considerations further justify modeling with a heavy-tailed distribution for the extreme event representation.
One obvious benefit is that it allows better model resolution along the representation axis, {\it i.e.}, better risk stratification.
For light-tail representations, extreme examples are clustered in a narrow region where the tail vanishes, thus a standard (light-tailed) learning model will report the average risk in that region.
However, if the representations are more spread out, then there is a more gradual change in risk, which can be better captured, as shown in Figure~\ref{fig:VItailestimation}.
Another argument for favoring heavy-tailed representations is that heavy-tailed phenomena are very common in nature \shortcite{bryson1974heavy}, and these tail samples are often encoded less robustly due to the lack of training examples.
Allowing long-tail representations relieves the burden of an encoder. 

%

\paragraph{Model structure.}
We consider latent variable model $p_\theta(y,x,z) = p_\theta(y|z)p_\theta(x|z)p(z)$, where $v=\{x,y\}$ are the observed variables.
Under the VI framework, similar to \eqref{eq:elbo} we write the $\ELBO(v;p_\theta(v,z),q_\phi(z|v))$ as
%
\begin{align}
& \EE_{Z\sim q_\phi(z|v)}[\log p_\theta(y|Z)] + \EE_{Z\sim q_\phi(z|v)}[\log p_\theta(x|Z)] \notag \\
& - \text{KL}(q_\phi(z|v)\parallel p(z)) \label{eq:elbonew}
\end{align}
%
where $p_\theta(y|z)$ is specified as an additive isotonic regression model, $p(z)$ is modeled with EVT, and the approximate posterior, $q_\phi(z|v)$, is specified as an inverse auto-regressive flow.
Note that unlike in the standard ELBO in \eqref{eq:elbonew}, we have dropped the term $\EE_{Z\sim q_\phi(z|v)}[\log p_\theta(x|z)]$ because we are not interested in modeling the covariates. Note this coincides with the {\it variational information bottleneck} (VIB) formulation \shortcite{alemi2016deep}. Additionally, the posterior $q_\phi(z|v)$ will not be conditioned on $y$, but only on $x$, because in practice, the labels $y$ are not available at inference time. 
%
%
Specifically, we rewrite the objective in \eqref{eq:elbonew} as
%
\begin{align}
     & \Psi_\beta(x,y;p_\theta(y|z),q_\phi(z|x)) = \label{eq:elbonewnew} \\
     & \hspace{10mm} \EE_{Z\sim q_\phi(z|x)}[\log p_\theta(y|Z)] - \beta \KL(q_\phi(z|x)\parallel p(z)) , \notag 
\end{align}
%
%
where $\beta$ is a hyperparameter controlling the relative contribution of the KL term to the objective.
Below we provide details for each component of the proposed approach.

\subsection{Decoder: Additive Monotonic Neural Network}
First, let us consider the following monotone mapping
    $\int_l^z h(s;\theta)ds + \gamma$,
consisting on integrating a non-negative function $h(s;\theta)$ specified as a neural network with one-dimensional input, $s$, and parameterized by $\theta$. 
The choice of the lower end $l$ is arbitrary, and $\gamma$ is a bias term.
For multi-dimensional latent representation $z\in \BR^p$, we write the additive monotonic neural network (AMNN) as
%
%
%
%
%
%
\beq\label{eq:monotoneFunc}
    H(z;\theta) = \sum_j^p [\alpha_j\int_l^{z_j} h_j(s;\theta)ds] + \gamma ,
\eeq
where $\alpha_j$ serves as a weight which controlling the effect directions. In other words, when $\alpha_j>0$, it can be interpreted as an event stimulator; otherwise it is an event blocker.
To ensure $h(s;\theta)$ is non-negative, we apply exponential activation function to the network's output.
The integration of $z$ is conducted with numerical integration by
the Riemann-Stieltjes method \shortcite{davis2007methods}. 
%

From \eqref{eq:monotoneFunc} we obtain $\log p_\theta(y|z)=\ell_{\text{CLL}}(y,H(z;\theta))$, where $\ell_{\text{CLL}}(y, a)= \log \{\mathbbm{1}_{y=1}(y)(1 - \exp(-\exp(a)))+\mathbbm{1}_{y=0}(y)\exp(-\exp(a))\}$ is the complementary log-log (CLL) link, where $\mathbbm{1}_{(\cdot)}$ is the indicator function.
We prefer CLL over the standard logistic link 
since the CLL link is more sensitive at the tail end \shortcite{aranda1981two}.
\subsection{Latent Prior: Gaussian GPD Mixture}
To better capture the tail behavior of the latent representation, we assume random variable $Z \sim p(z)$ is a mixture of a standard Gaussian distribution truncated at $u$ and a GPD for modeling the tail end thresholded at $u$, {\it i.e.}, $F(z) = \Phi(z)$ when $z\leq u$ and $F(z)=\Phi(u) + (1-\Phi(u)) G_{\xi,\sigma}(z-u)$ when $z>u$,
%
%
where $\Phi(z)$ denotes the CDF of a standard Gaussian distribution.
Note that for $z>u$, $F(z)$ can be expressed as a GPD with parameters $(\tilde{\xi}, \tilde{\sigma}, \tilde{u})$ \shortcite{mcneil1997estimating}, where $\tilde{\xi} = \xi$ and if $\xi \neq 0$, $\tilde{\sigma} = \sigma(1-\Phi(u))^{\xi}$ and $\tilde{u} = u - \tilde{\sigma}((1-\Phi(u))^{-\xi}-1)/\xi$.
%
Otherwise, when $\xi = 0$, $\tilde{\sigma} = \sigma$ and $\tilde{u} = u+\tilde{\sigma}\log(1-\Phi(u))$. 
Consequently, the CDF for the mixed GPD is given by
\beq\label{eq:mixedGPD}
\begin{aligned}
F(z)&=\mathbbm{1}_{(-\infty, u]}(z)\Phi(z) + \mathbbm{1}_{(u, \infty)}(z)G_{\tilde{u},\xi,\tilde{\sigma}}(z).
\end{aligned}
\eeq
%
For simplicity, we denote the set of parameters in GPD as
${\psi}$=$\{\xi_{\text{GPD}}$,$\sigma_{\text{GPD}}\}$ and the threshold $u$ is a user-defined parameter.
In the experiments we set $u$ to $\Phi^{-1}(0.99)$.


\subsection{Latent Posterior: Inverse Autoregressive Flow}\label{sec:IAF}
Considering we have adopted a long-tailed GPD prior, we seek a posterior approximation $q_{\phi}(z|x)$ that is: ($i$) a flexible parameterization to approximate arbitrary distributions; and ($ii$) with a tractable likelihood to be able to evaluate the $\KL(q_\phi(z|x)\parallel p(z))$ exactly.
We need ($i$) because the true posterior is likely to exhibit heavy-tailed behavior due to the extended coverage of the GPD prior, and ($ii$) is to ensure accurate and low-variance Monte Carlo estimation of the KL-divergence at the tail end of the prior.
These requirements invalidate some popular choices, {\it e.g.}, a standard Gaussian posterior is light-tailed, and the implicit neural-sampler-based posterior typical in the work of adversarial variational Bayes \shortcite{mescheder2017adversarial}, does not have a tractable likelihood. 


One model family satisfying the above two requirements is known as the generative flows \shortcite{rezende2015variational}, where simple invertible transformations with tractable $\log$ Jacobian determinants are stacked together, transforming a simple base distribution into a complex one, while still having closed-form expressions for the likelihood.
In this work, we consider the {\it inverse autoregressive flow} (IAF) model \shortcite{kingma2016improved}.
The flow chain is built as:
\beq\label{eq:iaf}
z_t = \mu_t + \sigma_t \odot z_{t-1}, \text{ for } 1 \leq t \leq T, 
\eeq
where $\mu_t\in\mathbb{R}^p$ and $\sigma_t\in\mathbb{R}^p$ are learnable parameters, $\odot$ denotes the element-wise product,
$z_0$ is typically drawn from a $p$-dimensional Gaussian distribution, $z_0 \sim \mathcal{N}(\mu_0, \text{Diag}(\sigma^2_0))$
where $\mu_0$ and $\sigma_0$ are obtained from an initial encoder defined by a neural network given input $x$ with parameter $\phi$.
A sample from the posterior $q_{\phi}(z|x)$ is given by $z_T$, obtained by ``flowing'' $z_0$ through \eqref{eq:iaf}.
Provided the Jacobians $\frac{d\mu_t}{dz_{t-1}}$ and $\frac{d\sigma_t}{dz_{t-1}}$ are strictly upper triangular \shortcite{papamakarios2017masked},
we obtain the following closed-form expression for the log posterior
\begin{align}
    \log q(z|x) & = \log q(z_0|x) - \sum_{t=1}^T\log \text{det}\left|\frac{dz_t}{dz_{t-1}}\right|\label{eq:IAFposterior} \\
    & = -\sum_{j=1}^p \left(\frac{1}{2}e_j^2 + \frac{1}{2}\log(2\pi)+\sum_{t=0}^T\log \sigma_{t,j}\right)\nonumber,
\end{align}
where $e_j=(x_j-\mu_{0,j})/\sigma_{0,j}$ for the $j$th dimension.
\begin{table*}[ht]
\centering
\caption{Ablation study of VIE with different combinations of architectures on realistic synthetic datasets with $1\%$ event rate. The oracle model has used the ground-truth model parameters to predict. }
\label{tab:ablation}
\resizebox{\textwidth}{!}{
\begin{tabular}{@{}ccccccccccc@{}}
\toprule
              &             &                    &           &             & \multicolumn{3}{c}{Average AUC (standard deviation)}                                                                                       & \multicolumn{3}{c}{Average AUPRC (standard deviation)}                                                                                     \\ \midrule
              & Prior       & Encoder            & Decoder   & Prior Match & n=5k                                    & n=10k                                   & n=20k                                   & n=5k                                    & n=10k                                   & n=20k                                   \\ \midrule
VAE           & Gaussian    & Gaussian           & MLP       & True        & 0.552 (0.092)                           & 0.682 (0.030)                           & 0.674 (0.020)                           & 0.026 (0.010)                           & 0.053 (0.010)                           & 0.061 (0.017)                           \\
VAE-GPD       & mixed GPD   & Gaussian           & AMNN      & False       & 0.569 (0.062)                           & 0.599 (0.010)                           & 0.653 (0.027)                           & 0.021 (0.003)                           & 0.027 (0.005)                           & 0.035 (0.013)                           \\
IAF-GPD       & mixed GPD   & IAF   & AMNN      & False        & 0.511(0.021)                            & 0.551 (0.018)                           & 0.665 (0.029)                           & 0.017(0.002)                            & 0.019 (0.002)                           & 0.025 (0.008)                           \\
Fenchel-GPD   & mixed GPD   & Implicit           & AMNN      & True        & 0.623 (0.036)                           & 0.668 (0.044)                           & 0.694 (0.021)                           & 0.037 (0.010)                           & 0.048 (0.013)                           & 0.062 (0.026)                           \\
[3pt]
VIE           & mixed GPD   & IAF  & AMNN      & True        & \textbf{0.684} (0.031) & \textbf{0.697} (0.036) & \textbf{0.701} (0.017) & \textbf{0.050} (0.009) & \textbf{0.061} (0.025) & \textbf{0.079} (0.025) \\\midrule
\multicolumn{4}{c}{Oracle (with $90\%$ confidence interval)} &             & \multicolumn{3}{c}{0.704 {[}0.662, 0.751{]}}                                                                                & \multicolumn{3}{c}{0.092 {[}0.058, 0.141{]}}                                                                                \\ \bottomrule
\end{tabular}
}
\end{table*}

\subsection{Posterior Match with Fenchel Mini-Max Learning}
We consider an additional modification that explicitly encourages the match of the aggregated posterior $q_{\phi}(z) = \int q_{\phi}(z|x) p_d(x) \ud x$ to the prior $p(z)$, which has been reported to be vastly successful at improving VAE learning \shortcite{mescheder2017adversarial}.
In our case, $q_{\phi}(z)$ does not have a closed-form expression for the likelihood ratio of the $\KL$ formulation, which motivates us to use a sample-based estimator.
We consider the mini-max $\KL$ estimator based on the Fenchel duality \shortcite{tao2019fenchel, dai2018coupled}.
Concretely, recall the $\KL$ can be expressed in its Fenchel dual form \footnote{We have removed the constant term for notational clarity.}
\begin{align}
& \Gamma(p, q_\phi, \nu) =\EE_{Z\sim q_\phi(z)}[\nu(Z)] - \EE_{Z'\sim p(z)}[\exp(\nu(Z'))] \notag \\
& \KL(q_\phi(z)\parallel p(z)) = \max_{\nu \in \mathcal{F}} \Gamma(p, q, \nu) , \label{eq:fenchel}
\end{align}
where $\nu(z)$ is commonly known as the critic function in the adversarial learning literature, and we maximize wrt $\nu(z)$ in the space of all functions $\mathcal{F}$, modeled with a deep neural network. 
We use \eqref{eq:elbonewnew} and \eqref{eq:fenchel} to derive an augmented $\ELBO$ that further penalizes the discrepancy between the aggregated posterior and the prior, {\it i.e.},  $\Psi_\beta(x,y;p_{\theta}(y|z), q_{\phi}(z|x)) - \lambda \KL(q_{\phi}(z)\parallel p(z))$,
%
%
where $\lambda$ is a regularization hyperparameter \shortcite{chen2018wiener}.
Solving for this objective results in the following mini-max game
\beq
\max_{\theta, \phi} \min_{\nu} \Psi_\beta(x,y;p_{\theta}(y|z), q_{\phi}(z|x)) - \lambda \Gamma(p_{\theta}, q_{\phi}, \nu) , 
\eeq
where $\beta$ and $\lambda$ are regularization hyperparameters.
In a similar vein to $\beta$-VAE and adversarial variational Bayes (AVB), our objective leverages $\beta$, $\lambda>0$ to balance the prediction accuracy and the complexity of the latent representation via KL regularization.
Further, from $\Psi_\beta(x,y;p_{\theta}(y|z), q_{\phi}(z,x))$ in \eqref{eq:elbonewnew}, note that the decoder $p_{\theta}(y|z)$ is obtained from the additive neural network in \eqref{eq:monotoneFunc}, $p_{\psi}(z)$ is the Gaussian GPD mixture with CDF in \eqref{eq:mixedGPD}, $q_{\phi}(z|x)$ is the autoregressive flow implied by \eqref{eq:iaf} and $\nu(z;\omega)$ is the critic function specified as a neural network and parameterized by $\omega$.

To avoid collapsing to suboptimal local minima, we train the encoder 
arm more frequently to compensate for the detrimental posterior lagging phenomenon \shortcite{he2019lagging}.
The pseudo-code for the proposed VIE is summarized in Algorithm~\ref{algo:VIE} and detailed architecture can be found in the SM.

\begin{algorithm}[t!]
\SetAlgoLined
{\bf Data:} $\mathcal{D}=(x,y)$. $x$: inputs, {$y$}: labels \\
{\bf Networks and parameters:} \texttt{Init-Encoder}({$x,\epsilon;\phi$}): Initial encoder network; \texttt{IAF}({$z;\phi$}): recursive autoregressive neural network; $\nu$({$z;\omega$}): critic neural network; \\ \texttt{AMNN}({$z;\theta$}): additive monotonic neural net; \\
prior: $p_{\psi}({z})=$\texttt{MixedGPD}($z$;${\psi}$,$u$), ${\psi}$=$\{\xi_{\text{GPD}}$,$\sigma_{\text{GPD}}\}$\\
%
{\bf Initialize:} \texttt{Init-Encoder}, \texttt{IAF}, $\nu$, \texttt{AMNN}, $\psi$\\
 \For{\text{iteration} $k \in \{1,\ldots,K\}$}{
      Sample $\{(x_i, y_i)\}_{i=1}^m$ from $\mathcal{D}$, {$\{\epsilon_i\}_{i=1}^m$} from $p(\epsilon)$\\
      {$[\mu_0,\sigma_0]$} =\texttt{Init-Encoder}({$x,\epsilon;\phi$})\\
          Sample ${z}_{\texttt{pr}}$ from $p_{\psi}({z})$, $\bm{z}_0$ from $\mathcal{N}({\mu}_0,{\Sigma}_0)$\\ 
          Compute  $l_{\texttt{post}}:=\log q_{\phi}(z_0|x)$\\
      \For{step $t \in \{1,\ldots,T\}$}{
      $[{\mu}_{t},{\sigma}_{t}]=$\texttt{IAF}(${z}_{t-1};\phi$), ${z}_t = {\mu}_{t} + {\sigma}_{t} \odot {z}_{t-1}$\\
      $l_{\texttt{post}}=l_{\texttt{post}}-\sum(\log {\sigma_{t}})$
      }
      $\log p_{\theta}(y|{z}_T) = \ell_{\text{CLL}}(y,\text{AMNN}(z_T;\theta))$ \\
      {\bf Descend} $\omega$ by $\nabla_{\omega}\frac{1}{m}\sum[\nu_\omega({z}_{\texttt{pr}})-\log \nu_\omega({z}_T)]$\\
      {\bf Ascend} $\Omega = \{\phi, \psi, \theta\}$ by\\
      $\nabla_{\Omega}\frac{1}{m}\sum[\log p_{\theta}(y|{z}_T)-\log \nu_\omega({z}_T)-\text{KL}]$, where $\text{KL} = l_{\texttt{post}} - \log p_{\psi}({z}_T)$\\
       }
      
 \caption{Variational Inference with Extremals.
 }
 \label{algo:VIE}
\end{algorithm}

\section{RELATED WORK}
{\bf Rare-event modeling with regression.} Initiated by \citet{king2001logistic}, the discussion on how to handle the unique challenges presented by rare-event data for regression models has attracted extensive research attention.
The statistical literature has mainly focused on bias correction for sampling \shortcite{fithian2014local} and estimation \shortcite{firth1993bias}, driven by theoretical considerations in maximum likelihood estimation.
However, their assumptions are often violated in the face of modern datasets \shortcite{sur2019modern}, characterized by high-dimensionality and complex interactions.
Our proposal approaches a solution from a representation learning perspective \shortcite{bengio2013representation}, by explicitly exploiting the statistical regularities of extreme values to better capture extreme representations associated with rare events.

{\bf Re-sampling and loss correction.} Applying statistical adjustments during model training is a straightforward solution to re-establish balance, but often associated with obvious caveats. For example, the popular down-sampling and up-sampling \shortcite{he2009learning} discard useful information or introduce artificial bias, exacerbating the chances of capturing spurious features that may harm generalization \shortcite{drummond2003c4, cao2019learning}, and their performance gains may be limited \shortcite{byrd2019effect}.
While traditionally tuned by trial and error, recent works have explored automated weight adjustments 
\shortcite{lin2017focal, zhang2020class}
, and principled loss correction that factored in class-size differences \shortcite{cui2019class}.
Our contribution is orthogonal to these developments and promises additional gains when used in synergy.

{\bf Transferring knowledge from the majority classes.} 
Adapting the knowledge learned from data-rich classes to their under-represented counterparts has shown success in few-shot learning, especially in the visual recognition field \shortcite{wang2017learning, chen2020supercharging}, and also in the clinical setting \shortcite{bohning2015meta}.
However, their success often critically depends on strong assumptions, the violation of which typically severely undermines performance \shortcite{wang2020generalizing}.
Related are the one-class classification (OCC) models \shortcite{tax2002one}, assuming stable patterns for the majority over the minority classes.
Our assumptions are weaker than those made in these model categories, and empirical results also suggest the proposed VIE works more favorably in practice (see experiments).
\section{EXPERIMENTS}
We carefully evaluate the proposed VIE on a diverse set of realistic synthetic data and real-world datasets with different degrees of imbalance.
Our implementation is based on PyTorch, and code to replicate our experiments are available from \url{https://github.com/ZidiXiu/VIE/}. We provide additional experiments and analyses in the SM. 

\begin{table*}[t!]
\caption{Average AUC and AUPRC from real-world datasets.
}
\label{tab:realresultssummary}
\resizebox{\textwidth}{!}{
\begin{tabular}{@{}c|cc|cccc|cc|c|cc|cccc|cc|c@{}}
\toprule
           & \multicolumn{9}{|c|}{average AUC}                                                                                                                        & \multicolumn{9}{|c}{average AUPRC}                                                                                                                      \\ \midrule
           & \multicolumn{2}{|c|}{COVID}        & \multicolumn{4}{|c|}{InP}                                           & \multicolumn{2}{|c|}{SEER}        & SLEEP          & \multicolumn{2}{|c|}{COVID}        & \multicolumn{4}{|c|}{InP}                                           & \multicolumn{2}{|c|}{SEER}        & SLEEP          \\\midrule
Event category & Mortality          & Combined            & 12\small{h}            & 24\small{h}            & 48\small{h}            & 168\small{h}            & 3\small{mo}            & 11\small{mo}           & 600\small{d}            & Mortality          & Combined            & 12\small{h}            & 24\small{h}            & 48\small{h}            & 168\small{h}            & 3\small{mo}            & 11\small{mo}           & 600\small{d} \\\midrule
LASSO      & 0.856          & 0.853          & 0.822          & 0.789          & 0.767          & 0.760          & 0.888          & 0.845          & 0.720          & 0.235          & \textbf{0.542} & 0.092          & 0.131          & 0.159          & 0.216          & 0.140          & 0.309          & 0.164          \\
MLP        & 0.862          & 0.854          & 0.824          & 0.806          & 0.762          & 0.768          & 0.885          & 0.856          & 0.730          & 0.225          & 0.531          & 0.093          & 0.141          & 0.159          & 0.221          & 0.169          & 0.322          & 0.182          \\
DeepSVDD   & NA             & NA             & 0.633          & 0.608          & 0.605          & 0.551          & 0.592          & 0.572          & 0.644          & NA             & NA             & 0.020          & 0.030          & 0.044          & 0.063          & 0.026          & 0.068          & 0.118          \\
IW         & 0.856          & 0.860          & 0.776          & 0.748          & 0.726          & 0.728          & 0.798          & 0.832          & 0.642          & 0.193          & 0.511          & 0.073          & 0.086          & 0.105          & 0.165          & 0.123          & 0.274          & 0.120          \\
Focal      & 0.829          & 0.854          & 0.750          & 0.779          & 0.741          & 0.705          & 0.868          & 0.835          & 0.633          & 0.238          & 0.484          & 0.044          & 0.112          & 0.120          & 0.149          & 0.141          & 0.263          & 0.101          \\
LDAM       & 0.857          & 0.843          & 0.819          & 0.805          & 0.785          & 0.774         & 0.893       &  0.861            & 0.755          & 0.202          & 0.535          & 0.086          & 0.130          & 0.148          & 0.197          & 0.177           & 0.332         & 0.179          \\
[5pt]
VIE      & \textbf{0.883} & \textbf{0.867} & \textbf{0.840} & \textbf{0.818} & \textbf{0.793} & \textbf{0.780} & \textbf{0.895} & \textbf{0.862} & \textbf{0.778} & \textbf{0.268} & 0.535          & \textbf{0.100} & \textbf{0.150} & \textbf{0.179} & \textbf{0.240} & \textbf{0.189} & \textbf{0.345} & \textbf{0.196} \\ \bottomrule
\end{tabular}
}
\end{table*}
%
{\bf Baseline Models}
We consider the following set of competing baselines to compare the proposed solution: 
LASSO regression \shortcite{tibshirani1996regression}, MLP with re-sampling and re-weighting (MLP), 
Importance-Weighting model (IW) \shortcite{byrd2019effect}, FOCAL loss \shortcite{lin2017focal}, 
Label-Distribution-Aware Margin loss (LDAM) \shortcite{cao2019learning}, 
and SVD based one-class classification model (Deep-SVDD) \shortcite{ruff2018deep}.
We tune the hyper-parameters of baseline models on the validation dataset, and pick best performing hyper-parameters to evaluate test set performance.
For detailed settings please refer to the SM.

{\bf Evaluation Metrics} To quantify model performance, we consider AUC and AUPRC.
AUC is the area under the Receiver Operating Characteristic (ROC) curve, which provides a threshold-free evaluation metric for classification model performance.
AUC summarizes the trade-off between True Positive Rate (TPR) and False Positive Rate (FPR).
AUPRC summarizes the trade-off between TPR and True Predictive Rate.
Specifically, it evaluates the area under Precision-Recall (PR) curve.
We discuss other metrics in the SM. 
In simulation studies, we repeat simulation ten times to obtain empirical AUC and AUPRC confidence intervals.
For real world datasets, we applied bootstrapping to estimate the confidence intervals.
\subsection*{Ablation study for VIE}
VIE applies a few state-of-art techniques in variational inference in order to achieve optimal performance. In this section, we decouple their contributions via an ablation study, to justify the necessity of including those techniques in our final model.
To this end, we synthesize a semi-synthetic dataset based on the Framingham study \shortcite{mitchell2010arterial}, a long-term cardiovascular survival cohort study. 
We use a realistic model to synthesize data from the real-world covariates under varying conditions, {\it i.e.}, different event rates, sample size, non-linearity, {\it etc.} 
More specifically, we use the CoxPH-Weibull model \shortcite{bender2005generating} to simulate the survival times of patients $T = \{\frac{-\log U}{\lambda \exp(g(x))}\}^{1/\nu}$, where $g(x)$ is either a linear function or a randomly initialized neural net.
Our goal is to predict whether the subject will decease within a pre-specified time frame, {\it i.e.}, $T < t_0$.
Via adjusting the cut-off threshold $t_0$, we can simulate different event rates. A detailed description of the simulation strategy is in the SM. 

We experiment with different combinations of advanced VI techniques, as summarized in Table~\ref{tab:ablation}. Limited by space, we report results at $1\%$ event rate with $g(\cdot)$ set to a randomly initialized neural network under various sample sizes.
Additional results on linear models and other synthetic datasets are consistent and can be found in the SM. 
IAF and GPD only variants perform poorly, even compared to the vanilla VAE solution. This is possibly due to the fact that priors are mismatched. Explicitly matching to the prior via Fenchel mini-max learning technique improves performance. However, without using an encoder with a tractable likelihood, the model cannot directly leverage knowledge from the GPD prior likelihood. Stacked together (mixed GPD+IAF+Fenchel), our full proposal of VIE consistently outperforms its variants, approaching oracle performance in the large sample regime. 
\subsection{Real-World Datasets}
To extensively evaluate real-world performance, we consider a wide range of real-world datasets, briefly summarized below: ($i$) \texttt{COVID}: A dataset of patients admitted to the DUHS with positive COVID-19 testing, to predict death or use of a ventilator. 
($ii$) \texttt{InP} \shortcite{o2020development}:
An in-patient data from DUHS, to predict the risk of death or ICR transfers. ($iii$) \texttt{SEER} \shortcite{ries2007seer}: A public dataset studying cancer survival among adults
curated by the U.S. Surveillance, Epidemiology, and End Results (SEER) Program, here we use a 10-year follow-up breast cancer subcohort.
($iv$) \texttt{SLEEP} \shortcite{quan1997sleep}: The Sleep Heart Health Study (SHHS) is a prospective cohort study about sleeping disorder and cardiovascular diseases. Summary statistics of these four real-world datasets are given in Table~\ref{tab:realsummary}. Note that \texttt{InP, SEER} and \texttt{SLEEP} are all survival datasets, among which \texttt{SEER} and \texttt{SLEEP} include censored subjects.
We  follow the
data pre-processing steps in \shortcite{xiu2020variational}. 
To create outcome labels, we set a cut-off time to define an event of interest the same as in the ablation study, and exclude subjects censored before the cut-off time.
The excluded samples only account for less than $0.2\%$ of the whole population, and therefore it is expected to have a very limited impact on our results. Datasets have been randomly split into training, validation, and testing datasets with ratio 6:2:2.
See the SM 
for details on data pre-processing.
%

\begin{table}[t!]
\centering
\caption{Summary statistics for real-world datasets.}
\resizebox{\linewidth}{!}{
\begin{tabular}{lrrrr}
\toprule
                  & \textsc{COVID}       & \textsc{InP}              & \textsc{SEER}              & \textsc{SLEEP}                  \\ \midrule
sample size               &  25,315     & 67,655                & 68,082                & 5026                \\
dimension & 1268(668)  &   73(39)             & 789(771)               & 206(162)        \\
event rate ($\%$) & $2.6\%$, $8\%$ & $1\sim 5\%$ & $1\sim 5\%$ & $5\%$\\
\bottomrule
\end{tabular}
}
\label{tab:realsummary}
\end{table}

\begin{figure}[ht]
    \centering
    \includegraphics[width=\linewidth]{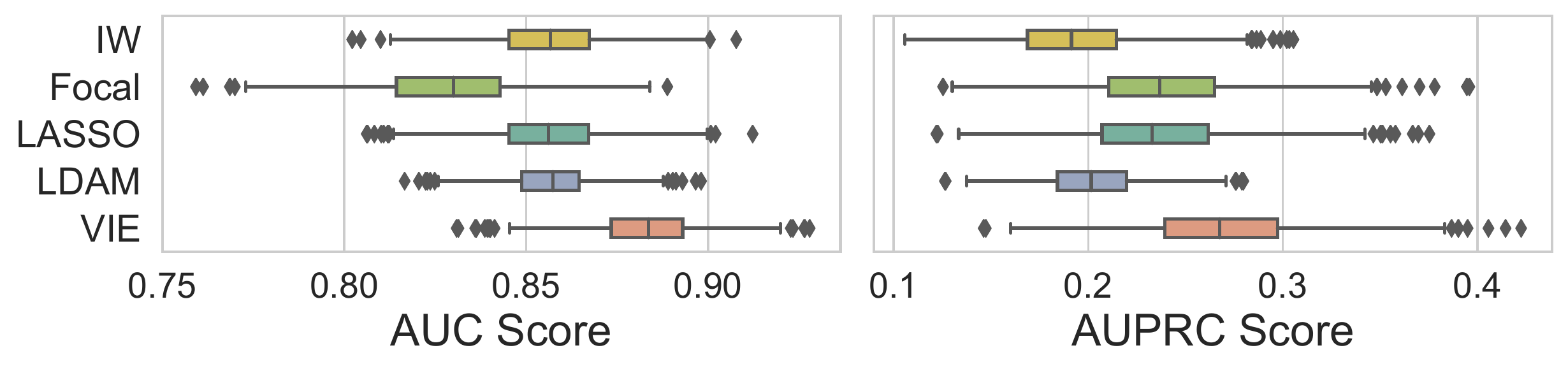}
\caption{Bootstrapped AUC (left) and AUPRC (right) distributions for the COVID mortality data (2.6\% event rate).}
\label{fig:COVID-mortality}
\end{figure}

Table~\ref{tab:realresultssummary} compares VIE to its counterparts, where the numbers are averaged over the bootstrap samples.
We see the proposed VIE yields the best performance in almost all cases, and the lead is more significant with low event rates.
Note that the one-class classification based DeepSVDD performs poorly, which implies treating rare events as outliers are inappropriate in the scenarios considered here.
Re-weighting and resampling based methods (IW, Focal) are less stable compared to those simple baselines (LASSO, MLP). 
The theoretically optimal LDAM works well in general, second only to VIE in most settings.
To further demonstrate the stability of our method, we visualize the bootstrapped evaluation scores for the COVID dataset in Figure~\ref{fig:COVID-mortality}, and defer the additional cross-validation results to the SM. 
We see that VIE leads consistently.

\begin{figure}[t!]
    \centering
    \includegraphics[width=\linewidth]{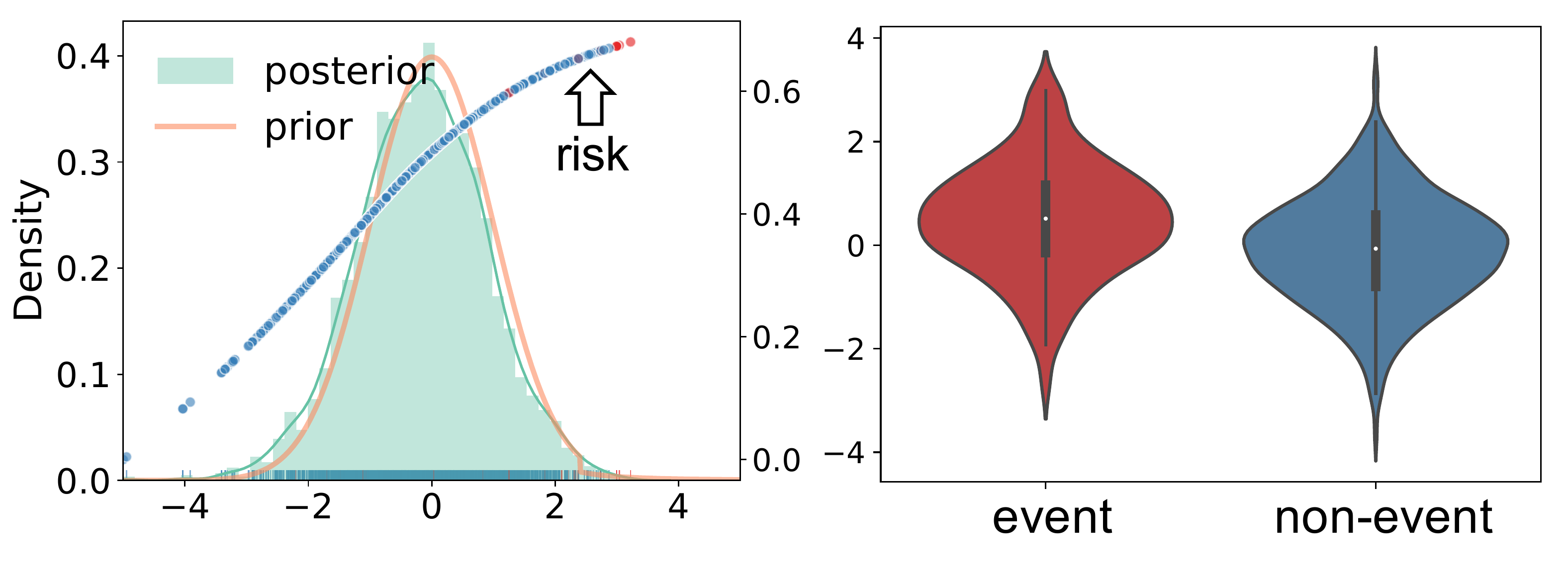}
    \caption{First latent dimension from the \texttt{InP} dataset (1\% event rate). Left: Learned prior and posterior distribution, and monotonic predicted risks (right axis). Right: The latent representation values distribution grouped by event type.}
    \label{fig:reallatent}
\end{figure}

We also verify empirically that the estimated GPD shape parameters  $\xi_{\text{GPD}}$ are mostly positive (see the SM), indicating heavier than Gaussian tails as we have hypothesized. In Figure~\ref{fig:reallatent}, we visualize one such latent dimension from the \texttt{InP} dataset, along with the associated risk learned by AMNN. In this example, the tail part is heavier than Gaussian and is associated with elevated risk. See our SM 
for examples where the extended tail contributes to prohibit the event. 


\section{CONCLUSIONS}
Motivated by the challenges of rare-event prediction in clinical settings, we presented Variational Inference with Extremals (VIE), a novel extreme representation learning-based variational solution to the problem.
In this model we leveraged GPD to learn the extreme distributions with few samples and applied additive monotonic neural networks to disentangle the latent dimensions' effects on the outcome. 
VIE featured better generalization and interpretability, as evidenced by a strong performance on real and synthetic datasets.
In future work, we will extend this framework to the context of causal inference to quantify treatment effects in the label imbalanced setting \shortcite{lu2020reconsidering}. 

\subsubsection*{Acknowledgements}
The authors would like to thank the Duke Institute for Health Innovation (DIHI) for providing access to curated COVID-19 data and outcomes.
This research was supported in part by NIDDK R01-DK123062 and NIH/NIBIB R01-EB025020.

{\small 
\bibliography{VIEVTref}
}


\beginsupplement
\include{SupplementalMaterial}
\end{document}

%% file: SupplementalMaterial.tex
\beginsupplement

\appendix
\onecolumn 


\section*{Supplementary Material to ``Variational Disentanglement for Rare Event Modeling''}
\tableofcontents
When the prevalence of an event is extremely low, but the event itself has substantial importance, the methods to identify such targets are called rare event modeling. Accurate and robust modeling of rare events is significant in many fields, such as identifying patients in high-risk and hopefully to prevent adverse outcomes from happening based on early intervention.

\begin{figure}[h]
    \centering
    \includegraphics[width=.4\linewidth]{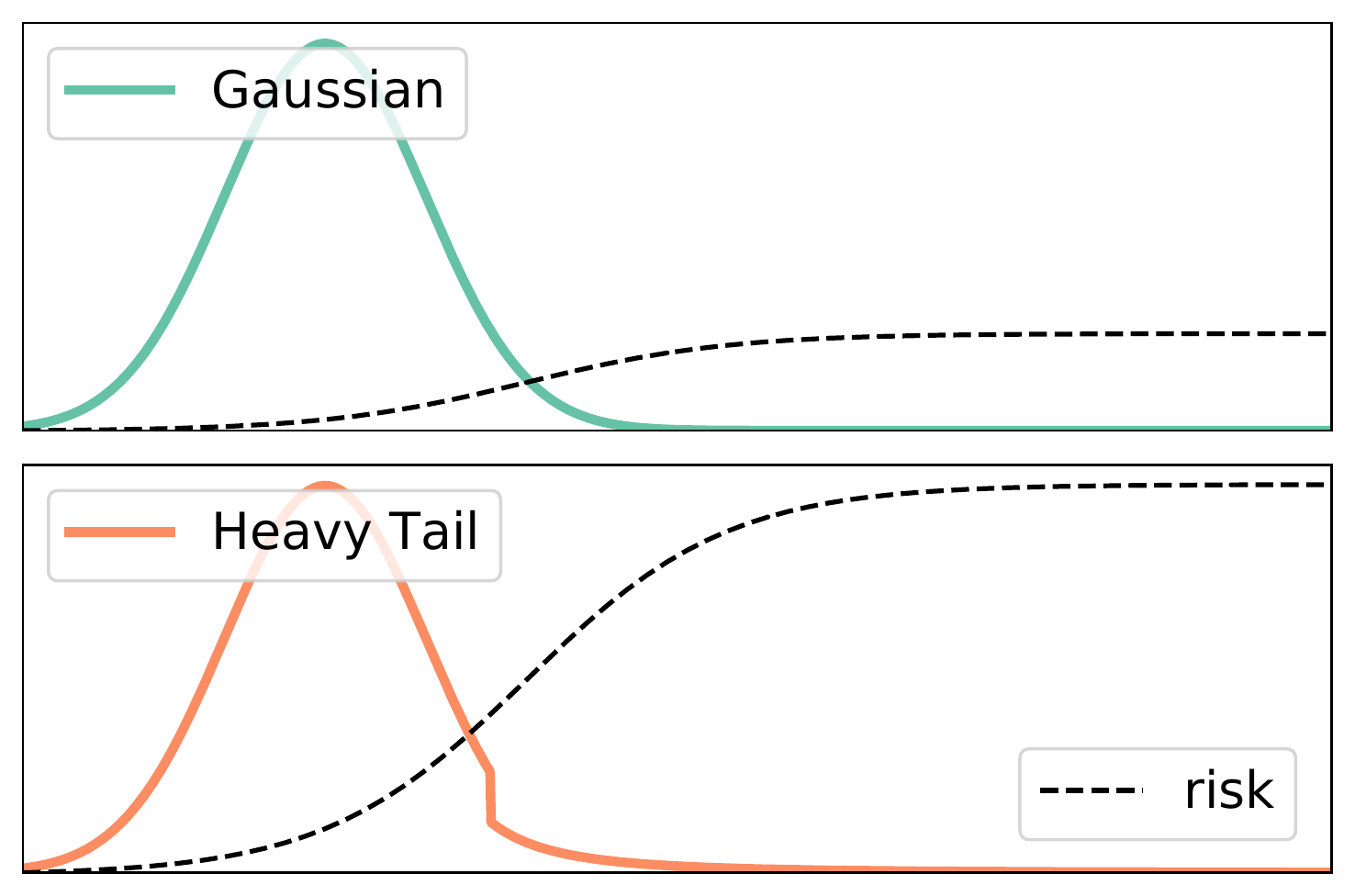}
    \caption{Feature representation mismatch at the tail parts. The heavy-tailed distribution can exploit extreme behavior in the latent space.}
    \label{fig:tail-risk}
\end{figure}

The scarcity of rare cases can cause extreme imbalanced among the dataset. Therefore, rare event modeling is challenging for most standard statistical approaches. As we discussed in the main text, careful statistical adjustments and new methodologies are required to approach such imbalance. Otherwise, the classifiers would be driven to the majority side and give misleading results. Also, the lack of representation in the minority class may cause unadjusted models to wrongly capturing spurious features that cannot generalize well to other observations. The apex of the risk curve or the mass of risk density usually overlays with the tail of the feature representation distribution, as illustrated in Figure~\ref{fig:tail-risk}, traditional statistical methods (such as Gaussian based approaches) often ill perform at the tail end, which can lead to lack-of-fit and poor generalization ability. 

We approach a solution to such challenges with a variational representation learning scheme that models disentangled extreme representations. Further, we design a robust, powerful prediction arm that combines the merits of a generalized additive model and isotonic neural net.

\subsection{A. Derivation of Mixed GPD tail distribution}\label{sup:GPD}
An important theory in {\it Extreme value theory} (EVT) shows that under some mild conditions, the conditional cumulative distribution of {\it exceedance} over a threshold $u$ follows Generalized Pareto Distribution, $\text{GPD}(u,\xi,\sigma)$ \citepSM{mcfadden1978modeling}, which has the cumulative distribution function (CDF) as:
\begin{equation*}
        G_{\xi,\sigma,u}(x)= 
\begin{cases}
    1-[1+\xi (x-u)/\sigma]^{-\frac{1}{\xi}},& \text{if }\xi \ne 0\\
    1-\exp[-(x-u)/\sigma],              & \text{if }\xi = 0
\end{cases}
\end{equation*}
where $\sigma$ is a positive scale parameter. According the shape parameter $\xi$, $x$ could have different support. When $\xi < 0$, the exceedance $x$ has bounded support $0\le x \le u-\sigma/\xi$, otherwise $x$ is bounded by $0$ on the left. $u$ is the location parameter. 
The corresponding PDF is:
\begin{equation*}
      g_{\xi,\sigma,u}(x)= 
\begin{cases}
    \sigma^{-1}[1+\xi (x-u)/\sigma]^{-\frac{1}{\xi}-1},& \text{if }\xi \ne 0\\
    \sigma^{-1}\exp[-(x-u)/\sigma],              & \text{if }\xi = 0
\end{cases} 
\end{equation*}
Thus the log-likelihood function is:
\begin{equation*}
        \log \text{likelihood}(x;\xi,\sigma,u)= 
\begin{cases}
    -\log\sigma - (\frac{1}{\xi}+1)\log[1+\xi (x-u)/\sigma],& \text{if }\xi \ne 0\\
    -\log\sigma -(x-u)/\sigma,              & \text{if }\xi = 0
\end{cases}
\end{equation*}

To enable modeling of the extreme representations, we adopt the Generalized Pareto Distribution as the tail part of our new variational prior, and the regular bulk representations $z\le u$ with a standard Gaussian distribution. Then mixed extreme tail distribution has the form \citepSM{mcneil1997estimating},
$$
F(z)=P(Z\le z) = P(Z\le u)+ (1-P(Z \le u))F_u(z-u)
$$
When $z>u$, the tail estimator is,
$$\hat{F}(z) = (1-F_n(u))G_{u, \xi,\sigma,u}(z) + F_n(u)$$
to approximate $F(z)$. 
Now we show that $\hat{F}(z)$ also has a GPD distribution with same $\xi$ and the following scale and location parameters,
\begin{equation*}
\begin{cases}
\tilde{\sigma} = \sigma(1-F_n(u))^{\xi},\tilde{u} = u - \tilde{\sigma}((1-F_n(u))^{-\xi}-1)/\xi,& \text{if }\xi \ne 0\\
\tilde{\sigma} = \sigma, \tilde{u} = u+\tilde{\sigma}\log(1-F_n(u)),              & \text{if }\xi = 0
\end{cases}    
\end{equation*}
When $\xi = 0$,
\begin{equation*}
    \begin{aligned}
    \hat{F}(z) &= (1-F_n(u))(1-\exp(-(x-u)/\sigma)) + F_n(u)\\
    &= 1- (1-F_n(u))\exp(-(x-u)/\sigma)\\
    &= 1 - \exp(\log(1-F_n(u)))\exp(-(x-u)/\sigma)\\
    &= 1 - \exp(-\frac{1}{\sigma}(x-u-\sigma\log(1-F_n(u))))\\
    &=1 - \exp(-\frac{1}{\tilde{\sigma}}(x-\tilde{u}))
    \end{aligned}
\end{equation*}
When $\xi \ne 0$,
\begin{equation*}
    \begin{aligned}
    \hat{F}(z) &= (1-F_n(u))( 1-(1+\xi (x-u)/\sigma)^{-\frac{1}{\xi}}) + F_n(u)\\
    &= 1- (1-F_n(u))(1+\xi (x-u)/\sigma)^{-\frac{1}{\xi}}\\
    &=  1- [(1-F_n(u))^{-\xi}(1+\xi (x-u)/\sigma)]^{-\frac{1}{\xi}}\\
    &= (1-F_n(u))^{-\xi} + (1-F_n(u))^{-\xi}\cdot \xi (x-u)/\sigma\\
    &=\frac{1}{(1-F_n(u))^{\xi}} + \frac{\xi (x-u)}{\sigma(1-F_n(u))^{\xi}}\\
    &=\frac{\sigma}{\tilde{\sigma}} + \frac{\xi (x-u)}{\tilde{\sigma}}\\
    &= 1 + \frac{\sigma - \tilde{\sigma}+\xi(x-\tilde{u})}{\tilde{\sigma}}\\
    &= 1 + \frac{\xi(x-\tilde{u} + \xi^{-1}\sigma - \xi^{-1}\tilde{\sigma})}{\tilde{\sigma}}\\
    \end{aligned}
\end{equation*}
Therefore, $\tilde{u} = \tilde{u} - \xi^{-1}\sigma - \xi^{-1}\tilde{\sigma}$.

\subsection{B. Implementation Details}\label{sup:implementation}
Our main algorithm was written in PyTorch (version 1.3.1) \citepSM{paszke2017automatic}. The experiments were conducted on an Intel(R) Xeon(R) and Tesla P100-PCIE-16GB GPU (except for the \texttt{COVID} dataset). The \texttt{COVID} dataset were stored and analyzed on a protected virtual network space with Inter(R) Xeon(R) Gold 6152 CPU 2.10GHz 2 Core(s). 
\subsubsection{Model Structure.}
In VIE, we end up optimizing the following objective,
%
\beq
\max_{\theta, \phi} \min_{\nu} \{ \EE_{x,y \sim \mathcal{D}}[\Psi_{\beta}(x,y; p_{\theta}, q_{\phi}) - \lambda \Gamma(p_{\theta}, q_{\phi}, \nu)] \} ,
\eeq
where
\begin{align*}
\Psi_{\beta}(x,y; p_{\theta}, q_{\phi}) = & \ \EE_{Z\sim q_{\phi}(z|x)}[\log p_{\theta}(y|Z)] \\ 
& - \hspace{6mm} \beta \text{KL}[q_{\phi}(z|x)||p(z)] ,
\end{align*}
Note that the GPD parameters $(\xi, \sigma)$ are absorbed in $\phi$, and hyperparameter $u$ is used in the GPD prior $p(z)$. $u$ is set to be $F^{-1}_z(0.99)$ in all experiments. When the event rate is $\ge 1\%$, we set $\lambda, \beta = (\num{1e-3}, \num{1e-5})$, otherwise we shrink the parameters to $\lambda, \beta = (\num{1e-4}, \num{1e-6})$.

More concretely, the constituting parts $p_{\theta}(y|z)$, $p_{\theta}(z|x)$, $q_{\phi}(z|x)$ and $\nu(z)$ are specified as follows
%
\begin{align}
    \begin{aligned}
        p_{\theta}(y|z) & \leftarrow \Phi(H( z;\theta)) \text{ Log-Log link } (\ref{eq:loglog}), \\
        H( z;\theta) & \leftarrow \text{ Additive Monotone Neural Net (\ref{eq:monotoneFunc-aprox}) with }\\
        p(z) & \leftarrow \text{Mixed GPD }(u, \xi_p, \sigma_p)\,\, (\ref{eq:mixedGPD}),  p=4\\
        q_{\phi}(z|x) & \leftarrow \text{Inverse Autoregressive Flow } (\ref{eq:IAFposterior}), \text{nstep}=5\\
        \nu(z) & \leftarrow \text{Standard neural network}.
    \end{aligned}
\end{align}
\begin{algorithm}[t!]
\SetAlgoLined
{\bf Data:} $\mathcal{D}=(x,y)$. $x$: inputs, {$y$}: labels \\
{\bf Networks and parameters:} \texttt{Init-Encoder}({$x,\epsilon;\phi$}): Initial encoder network; \texttt{IAF}({$z;\phi$}): recursive autoregressive neural network; $\nu$({$z;\omega$}): critic neural network; \\ \texttt{AMNN}({$z;\theta$}): additive monotonic neural net; \\
prior: $p_{\psi}({z})=$\texttt{MixedGPD}($z$;${\psi}$,$u$), ${\psi}$=$\{\xi_{\text{GPD}}$,$\sigma_{\text{GPD}}\}$\\
%
{\bf Initialize:} \texttt{Init-Encoder}, \texttt{IAF}, $\nu$, \texttt{AMNN}, $\psi$\\
 \For{\text{iteration} $k \in \{1,\ldots,K\}$}{
      Sample $\{(x_i, y_i)\}_{i=1}^m$ from $\mathcal{D}$, {$\{\epsilon_i\}_{i=1}^m$} from $p(\epsilon)$\\
      {$[\mu_0,\sigma_0]$} =\texttt{Init-Encoder}({$x,\epsilon;\phi$})\\
          Sample ${z}_{\texttt{pr}}$ from $p_{\psi}({z})$, $\bm{z}_0$ from $\mathcal{N}({\mu}_0,{\Sigma}_0)$\\ 
          Compute  $l_{\texttt{post}}:=\log q_{\phi}(z_0|x)$\\
      \For{step $t \in \{1,\ldots,T\}$}{
      $[{\mu}_{t},{\sigma}_{t}]=$\texttt{IAF}(${z}_{t-1};\phi$), ${z}_t = {\mu}_{t} + {\sigma}_{t} \odot {z}_{t-1}$\\
      $l_{\texttt{post}}=l_{\texttt{post}}-\sum(\log {\sigma_{t}})$
      }
      $\log p_{\theta}(y|{z}_T) = \ell_{\text{CLL}}(y,\text{AMNN}(z_T;\theta))$ \\
      {\bf Descend} $\omega$ by $\nabla_{\omega}\frac{1}{m}\sum[\nu_\omega({z}_{\texttt{pr}})-\log \nu_\omega({z}_T)]$\\
      {\bf Ascend} $\Omega = \{\phi, \psi, \theta\}$ by\\
      $\nabla_{\Omega}\frac{1}{m}\sum[\log p_{\theta}(y|{z}_T)-\log \nu_\omega({z}_T)-\text{KL}]$, where $\text{KL} = l_{\texttt{post}} - \log p_{\psi}({z}_T)$\\
       }
      
 \caption{Variational Inference with Extremals.
 }
 \label{algo:VIE-full}
\end{algorithm}

Pseudo-code for VIE is presented in Algorithm~\ref{algo:VIE-full}. In all experiments, AMNN, IAF, $\nu(z)$ are specified in terms of two-layer MLPs of 32 hidden units with Rectified Linear Unit (ReLU) activation functions. The initial encoder \texttt{Init-Encoder} is specified as a three-layers MLPs of 32 hidden units. We set the minibatch size to $m=200$. The critic network $\nu(x)$ uses the RMSprop optimizer with learning rate $\num{1e-3}$, other parts of the algorithm used Adam optimizer with learning rate $\num{1e-4}$. To avoid over-fitting, we set the dimension of latent space as $4$ in all experiments.

Note we have used the {\it Complementary Log-Log} (CLL) link function for $p_{\theta}(y|z)$ in (\ref{eq:loglog}),
\beq\label{eq:loglog}
\Phi(a) = 1 - \exp(-\exp(a)),
\eeq
for the outcome model as opposed to the standard {\it Logistic} link $1/(1+\exp(-a))$. The CLL link is more sensitive at the tail end, so it is more frequently used in statistical models dealing with vanishing probabilities \citep{aranda1981two}. 

To avoid collapsing to suboptimal local minimums, we train the encoder arm more frequently to compensate for the detrimental posterior lagging phenomenon \citep{he2019lagging}. Our pseudo-code for VIE is summarized in Algorithm~\ref{algo:VIE}.

\subsubsection{Numerical Integration.}
Following \eqref{eq:monotoneFunc}, we divide region $[l, z_j]$ evenly into $M$ bins of width $d_j = \frac{z_j-l}{M}$, with $z_{j,M} = z_j$. For the $M$ bins, we select a random point $z_{j,k}^r$ in each bin. The integral approximation on support $[z_{j,k}, z_{j,k+1}]$ is the rectangular area $h_j(z_{j,k}^{(r)})*d_j$.
As a result, the integral $\int_0^{z_j} h(s;\theta)ds$ is approximated with $\sum_k^{M-1} h_j(z_{j,k}^{(r)})d_j$.
With this approximation \eqref{eq:monotoneFunc} can be written as:
\beq\label{eq:monotoneFunc-aprox}
H(z;\theta) = \sum_j^p \alpha_j d_j \sum_k^{M} h_j(z_{j,k}^{(r)}) + \gamma .
\eeq
We set $M=100$ and $l=-5$ in all the experiments.
\subsubsection{Discussions on Evaluation Metrics.} In the main text, we reported AUC and AUPRC instead of single evaluation metrics,  {\it e.g.}, overall accuracy or error rate. Standard statistical metrics like Brier Scores (BS) and Binary classification entropy (BCE) could be deceptive when the event rate is low, {\it e.g.}, $\le 10\%$ 
\citepSM{schmid2014multivariate}. We will add BCE loss and the positive case BCE loss in the following sections in the simulation study for reference. Some poorly performed models can have relatively low BCE scores. In this case, the ground truth (Oracle) is the best reference we have. 

\subsection{C. Ablation Study}\label{sup:simulation}
We examine model performance on two simulation strategies. The first one is the semi-synthetic dataset, which exploits the real-world covariates structures. The second one is a synthetic dataset based on our extreme representation assumptions.
\subsubsection{Semi-synthetic Datasets} 
We synthesize a semi-synthetic dataset based on the Framingham study \citep{mitchell2010arterial}, a long-term cardiovascular survival cohort study. 
After quality control, $40,046$ subjects with nine covariates (four continuous and five categorical) are included.

We use a realistic model to synthesize data from the real-world covariates under varying conditions, {\it i.e.}, different event rates, sample size, nonlinearity, {\it etc.} 
More specifically, we use the coxPH-Weibull model \citep{bender2005generating} to simulate the survival time of patients $T = \{\frac{-\log U}{\lambda \exp(g(x))}\}^{1/\nu}$, where $g(x)$ is either a linear function or a randomly initialized neural net.
Our goal is to predict whether the subject will decease within a pre-specified time frame, {\it i.e.}, $T < t_0$.
Via adjusting the cut-off threshold $t_0$, we can simulate different event rates. The details are shown in Algorithm~\ref{algo:simu1}.
\begin{algorithm}[h]
\SetAlgoLined
Extract covariates from \texttt{Framingham} Dataset \;
Set $\nu, \lambda$ (the parameters of cox-Weibull distribution)\;
Set time cut-bound $t_0$\;
Decide $g(x):\mathbb{R}^q \rightarrow \mathbb{R}$ form\;
\For{$i \in \{1,\ldots,n\}$}{
     Sample $u_i$ from \texttt{Unif}$(0,1)$\;
     Compute $t_i=\{\frac{-\log u_i}{\lambda \exp(g(x_i))}\}^{1/\nu}$\;
     Compute $y_i = \mathbbm{1}[t_i < t_0]$\;
     $d_i=(y_i,x_i)$\;
     }
\Return $\mathcal{D}=\{d_i;i=1\ldots n\}$
\caption{Semi-synthetic Data}
\label{algo:simu1}
\end{algorithm}

In our experiments, the performances when $g(\cdot)$ set as a randomized neural network or a linear function do not differ very much. For simplicity, we will present the results under the neural network settings. Apart from the results at $1\%$ event rate given in the main text, we will show the results at $0.5\%$ and $5\%$ event rates here. The oracle results are calculated with plugging in the true $g(x_i)$ in Algorithm~\ref{algo:simu1}, and the randomness is from the generating scheme of the survival time $t$.
\subsubsection{Additional Results for semi-synthetic datasets}
\begin{figure}[h!]
\centering
\begin{minipage}{0.6\linewidth}
 \centering
    \includegraphics[width=\linewidth]{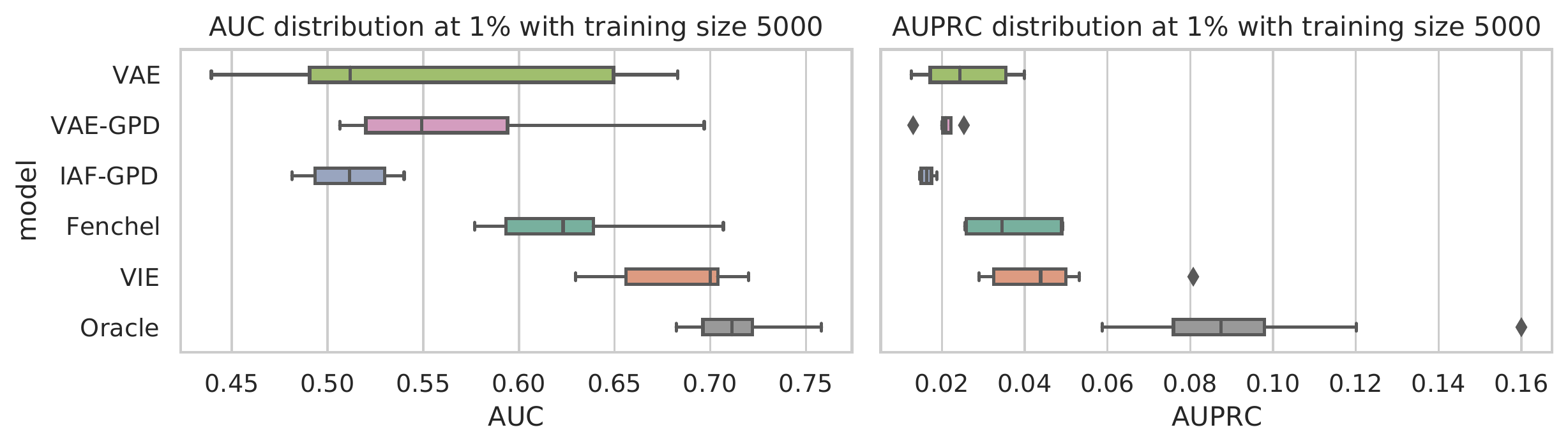}
 \end{minipage}
 \begin{minipage}{0.6\linewidth}
 \centering
    \includegraphics[width=\linewidth]{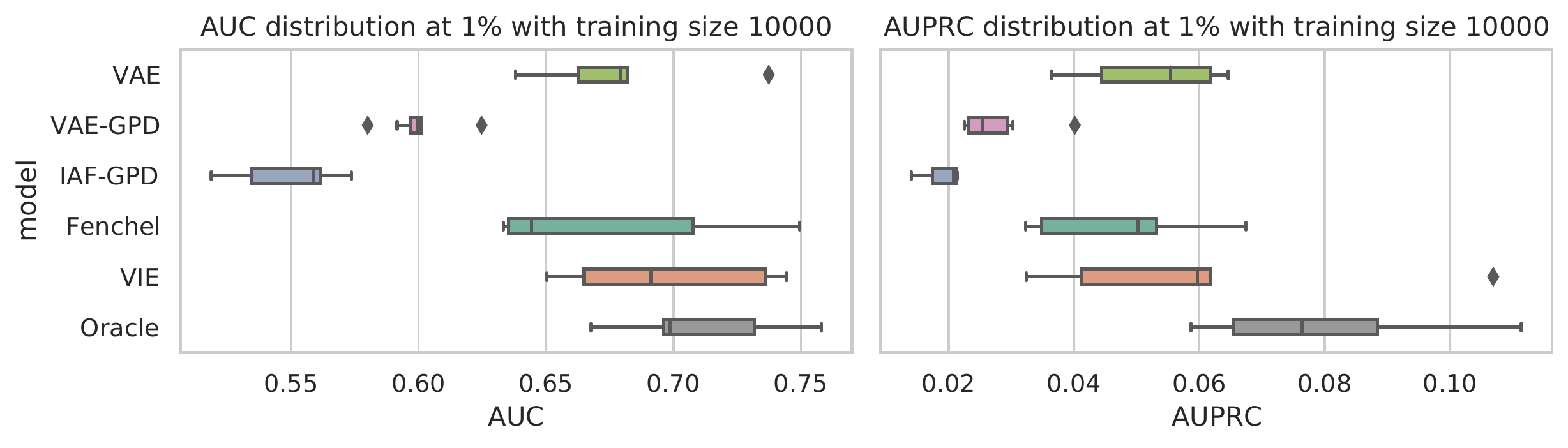}
 \end{minipage}
 \begin{minipage}{0.6\linewidth}
 \centering
    \includegraphics[width=\linewidth]{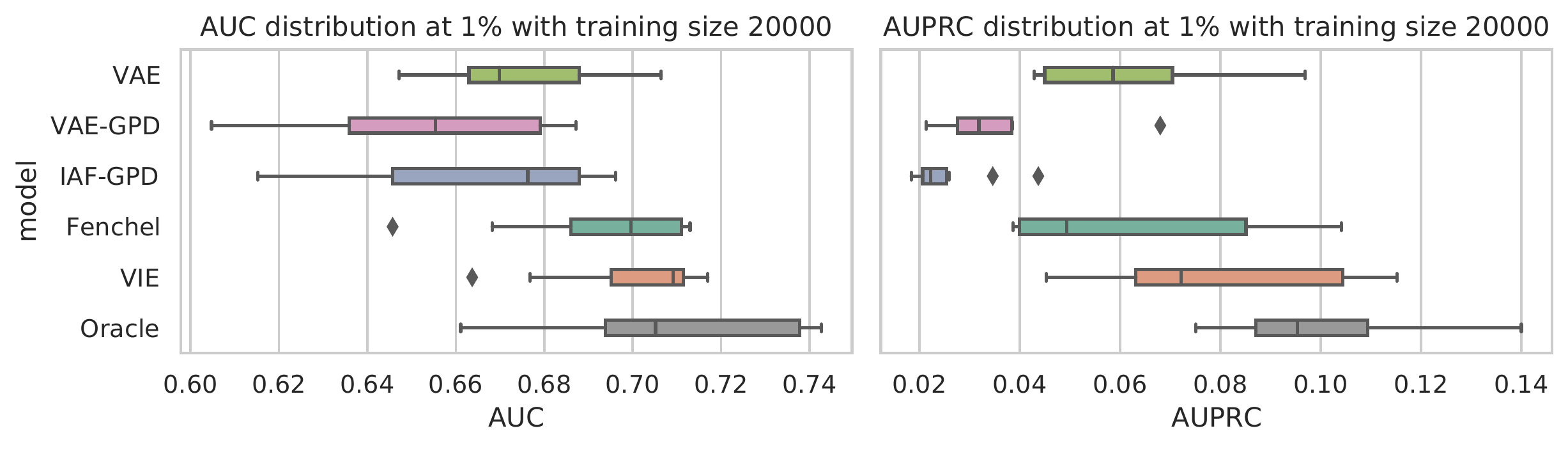}
 \end{minipage}
 \caption{Box plot of 10 independent $1\%$ event rate semi-synthetic analysis.}
 \label{fig:ablationer01}
\end{figure}

In $1\%$ event rate case presented in the main text, the AUC and AUPRC distributions are summarized in Figure~\ref{fig:ablationer01}, which corresponds to the average and standard deviation values presented in Table~\ref{tab:ablation}. 
We further examine the cases with $0.5\%$ and $5\%$ event rates to evaluate our method's robustness. Results are summarized in Table~\ref{tab:ablationer005} and Table~\ref{tab:ablationer05} respectively. Apart from the threshold-free metrics AUC and AUPRC, we also presented Binary Cross-Entropy loss (BCE) and the BCE loss associated with true events (positive case losses). Note that for an imbalanced dataset, BCE loss can be misleading. In the model, VAE-GPD, which is poorly-behaved in AUC and AUPRC, can have relatively low BCE loss since the majority group overwhelms the minority ({\it more important}) group. We can refer to the BCE loss and positive case loss in the oracle results for reference. VIE performs consistently close to the oracle results, especially with low event rate and small training sample size, and Fenchel-GPD is in the second place.  
\begin{table}[h]
\centering
\caption{Ablation study of VIE with $0.5\%$ event rate in semi-synthetic settings.}
\label{tab:ablationer005}
\resizebox{\textwidth}{!}{\begin{tabular}{cccccccllllll}
\hline
            & \multicolumn{3}{c}{Average AUC (std) $\uparrow$}         & \multicolumn{3}{c}{Average AUPRC (std)$\uparrow$}       & \multicolumn{3}{c}{Average BCE Loss (std) $\downarrow$}                                       & \multicolumn{3}{c}{Average Positive Case Loss (std) $\downarrow$}                             \\ \hline
            & n=5k          & n=10k         & n=20k         & n=5k          & n=10k         & n=20k         & \multicolumn{1}{c}{n=5k} & \multicolumn{1}{c}{n=10k} & \multicolumn{1}{c}{n=20k} & \multicolumn{1}{c}{n=5k} & \multicolumn{1}{c}{n=10k} & \multicolumn{1}{c}{n=20k} \\ \hline
VAE         & 0.494 (0.111) & 0.623 (0.102) & 0.697 (0.061) & 0.007 (0.005) & 0.017 (0.010) & 0.020 (0.007) & 0.498 (0.309)            & \textbf{0.035} (0.005)             & \textbf{0.031} (0.004)             & \textbf{0.010} (0.010)            & 0.027 (0.005)             & 0.026 (0.005)             \\
VAE-GPD     & 0.560 (0.045) & 0.602 (0.044) & 0.635 (0.045) & 0.008 (0.002) & 0.013 (0.010) & 0.016 (0.003) & 5.250 (2.096)            & 0.768 (0.258)             & 0.152 (0.222)             & 0.000 (0.000)            & \textbf{0.004} (0.001)             & \textbf{0.018} (0.008)             \\
IAF-GPD     & 0.631 (0.039) & 0.555 (0.042) & 0.533 (0.061) & 0.011 (0.007) & 0.008 (0.002) & 0.017 (0.013) & \textbf{0.032} (0.003)            & 0.038 (0.008)             & 0.043 (0.014)             & 0.027 (0.003)            & 0.024 (0.001)             & 0.027 (0.007)             \\
Fenchel-GPD & 0.615 (0.059) & 0.652 (0.055) & 0.667 (0.024) & 0.022 (0.016) & 0.021 (0.012) & \textbf{0.025} (0.008) & 0.034 (0.004)            & 0.037 (0.006)             & 0.033 (0.002)             & 0.028 (0.005)            & 0.032 (0.006)             & 0.027 (0.002)             \\
VIE         & \textbf{0.654} (0.074) & \textbf{0.692} (0.076) & \textbf{0.693} (0.036) & \textbf{0.022} (0.010) & \textbf{0.027} (0.015) & 0.024 (0.009) & 0.041 (0.018)            & 0.036 (0.003)             & 0.032 (0.003)             & 0.026 (0.005)            & 0.030 (0.005)             & 0.026 (0.003)             \\ \hline
Oracle      & \multicolumn{3}{c}{0.688 (0.618, 0.769)}      & \multicolumn{3}{c}{0.043 (0.023, 0.071)}      & \multicolumn{3}{c}{0.034 (0.028, 0.040)}                                         & \multicolumn{3}{c}{0.029 (0.023, 0.035)}                                         \\ \hline
\end{tabular}}
\end{table}
\begin{table}[h]
\centering
\caption{Ablation study of VIE with $5\%$ event rate in semi-synthetic settings.}
\label{tab:ablationer05}
\resizebox{\textwidth}{!}{\begin{tabular}{cccccccclllll}
\hline
           & \multicolumn{3}{c}{Average AUC (std) $\uparrow$}         & \multicolumn{3}{c}{Average AUPRC (std)$\uparrow$}       & \multicolumn{3}{c}{Average BCE Loss (std) $\downarrow$}                                       & \multicolumn{3}{c}{Average Positive Case Loss (std) $\downarrow$}                             \\ \hline
            & n=5k          & n=10k         & n=20k         & n=5k          & n=10k         & n=20k         & n=5k          & \multicolumn{1}{c}{n=10k} & \multicolumn{1}{c}{n=20k} & \multicolumn{1}{c}{n=5k} & \multicolumn{1}{c}{n=10k} & \multicolumn{1}{c}{n=20k} \\ \hline
VAE         & 0.594 (0.118) & 0.666 (0.021) & 0.693 (0.011) & 0.113 (0.049) & 0.144 (0.017) & 0.179 (0.018) & 0.308 (0.166) & 0.198 (0.004)             & 0.198 (0.009)             & 0.111 (0.046)            & 0.147 (0.003)             & 0.149 (0.008)             \\
VAE-GPD     & 0.583 (0.027) & 0.607 (0.014) & 0.663 (0.009) & 0.075 (0.008) & 0.087 (0.013) & 0.137 (0.024) & 2.106 (1.553) & 0.581 (0.056)             & 0.195 (0.011)             & \textbf{0.017} (0.015)            & \textbf{0.043} (0.004)             & 0.143 (0.013)             \\
IAF-GPD     & 0.664 (0.017) & 0.554 (0.033) & 0.503 (0.020) & 0.113 (0.022) & 0.063 (0.007) & 0.057 (0.003) & 0.194 (0.008) & 0.208 (0.004)             & 0.214 (0.009)             & 0.144 (0.009)            & 0.157 (0.004)             & 0.160 (0.008)             \\
Fenchel-GPD & 0.666 (0.014) & 0.687 (0.016) & 0.681 (0.008) & \textbf{0.145} (0.011) & \textbf{0.184 (0.022)} & 0.166 (0.013) & 0.196 (0.008) & \textbf{0.189} (0.007)             & 0.190 (0.011)             & 0.148 (0.008)            & 0.141 (0.007)             & 0.141 (0.012)             \\
VIE         & \textbf{0.679} (0.018) & \textbf{0.693} (0.027) & \textbf{0.693} (0.015) & 0.142 (0.018) & 0.172 (0.032) & \textbf{0.193}(0.013) & \textbf{0.188} (0.011) & 0.193 (0.005)             & \textbf{0.190} (0.006)             & 0.139 (0.013)            & 0.145 (0.006)             & \textbf{0.139} (0.008)             \\ \hline
Oracle      & \multicolumn{3}{c}{0.694 (0.670, 0.717)}      & \multicolumn{3}{c}{0.197 (0.179, 0.218)}      & \multicolumn{3}{c}{0.185 (0.179, 0.198)}                              & \multicolumn{3}{c}{0.137 (0.130, 0.149)}                                         \\ \hline
\end{tabular}}
\end{table}
\subsubsection{Long-tailed Synthetic Datasets}
We design the long-tailed synthetic datasets based on our proposed method, where the latent variable $z$ enjoys a long-tailed distribution. The pseudo-code for this simulation strategy is shown in Algorithm~\ref{algo:simu0}, where $t_0$ is a pre-specified time-cut, and $H(\cdot)$ is a randomized monotone neural network to create a monotone mapping from $z$ to the risk. 
\begin{algorithm}[!htbp]
\SetAlgoLined
Set sample size $n$, latent space dimension $p$, number of covariates $q$ \;
Set $\mu_p, \Sigma_p, \xi_p, \sigma_p$ (the parameters of a long-tailed distribution)\;
Set $\nu, \lambda$ (the parameters of cox-Weibull distribution)\;
Set time cut-bound $t_0$\;
Initialize $\psi$ (for MLP $g(\cdot;\psi): \mathbb{R}^p \rightarrow \mathbb{R}^q$), $\theta$ (for AMNN $H(\cdot;\theta): \mathbb{R}^p \rightarrow \mathbb{R}$)\;
\For{$i \in \{1,\ldots,n\}$}{
     Sample $z_i$ from $\text{ mixed GPD }(\mu_p, \Sigma_p, \xi_p, \sigma_p)$\;
     Compute $x_i=g(z_i;\psi)$\;
     Sample $u_i$ from \texttt{Unif}$(0,1)$\;
     Compute $t_i=\{\frac{-\log u_i}{\lambda \exp(H(z_i;\theta))}\}^{1/\nu}$\;
     Compute $y_i = I(t_i < t_0)$\;
     $d_i=(y_i,x_i)$
     }\;
\Return $D=\{d_i;i=1\ldots n\}$
\caption{Generation of long-tailed data.}
\label{algo:simu0}
\end{algorithm}

Table~\ref{tab:ablationersyn} summarizes the findings with the long-tailed distributed latent space datasets, with $1\%$ event rate. VIE can achieve relatively high AUC and AUPRC even with a small training sample size, which suggests that the proposed method can recover the long-tailed behavior in the feature representation. Among the combinations of different VI techniques, the Fenchel duality mechanism facilitates the distribution matching the best among other inference techniques.
\begin{table}[!htbp]
\centering
\caption{Ablation study of VIE with $1\%$ event rate in longtailed-synthetic settings}
\label{tab:ablationersyn}
\resizebox{\textwidth}{!}{\begin{tabular}{cccccccclllll}
\hline
           & \multicolumn{3}{c}{Average AUC (std) $\uparrow$}         & \multicolumn{3}{c}{Average AUPRC (std)$\uparrow$}       & \multicolumn{3}{c}{Average BCE Loss (std) $\downarrow$}                                       & \multicolumn{3}{c}{Average Positive Case Loss (std) $\downarrow$}                             \\ \hline
            & n=5k          & n=10k         & n=20k         & n=5k          & n=10k         & n=20k         & n=5k           & \multicolumn{1}{c}{n=10k} & \multicolumn{1}{c}{n=20k} & \multicolumn{1}{c}{n=5k} & \multicolumn{1}{c}{n=10k} & \multicolumn{1}{c}{n=20k} \\ \hline
VAE         & 0.722 (0.140) & 0.741 (0.099) & 0.798 (0.034) & 0.128 (0.076) & 0.119 (0.064) & 0.177 (0.034) & 0.138 (0.213)  & 0.121 (0.200)             & 0.055 (0.004)             & 0.029 (0.010)            & 0.034 (0.009)             & 0.039 (0.005)             \\
VAE-GPD     & 0.498 (0.039) & 0.450 (0.021) & 0.441 (0.055) & 0.013 (0.002) & 0.009 (0.000) & 0.009 (0.002) & 12.097 (5.582) & 22.445 (10.247)           & 13.438 (7.159)            & 0.000 (0.000)            & 0.000 (0.000)             & 0.001 (0.002)             \\
IAF-GPD     & 0.688 (0.021) & 0.632 (0.037) & 0.555 (0.039) & 0.097 (0.019) & 0.078 (0.027) & 0.046 (0.014) & 0.051 (0.005)  & 0.055 (0.003)             & 0.062 (0.007)             & 0.040 (0.005)            & 0.044 (0.004)             & 0.051 (0.007)             \\
Fenchel-GPD & 0.804 (0.028) & 0.807 (0.026) & 0.818 (0.020) & 0.174 (0.030) & 0.155 (0.052) & 0.166 (0.044) & 0.054 (0.007)  & 0.051 (0.003)             & \textbf{0.047} (0.004)             & 0.041 (0.007)            & 0.040 (0.004)             & 0.037 (0.005)             \\
VIE         & \textbf{0.823} (0.024) & \textbf{0.810} (0.023) & \textbf{0.836} (0.026) & \textbf{0.175} (0.036) & \textbf{0.163} (0.044) & \textbf{0.202} (0.024) & \textbf{0.047} (0.005)  & \textbf{0.050} (0.003)             & 0.049 (0.005)             & \textbf{0.037} (0.004)            & \textbf{0.040} (0.003)             & \textbf{0.037} (0.005)             \\ \hline
Oracle      & \multicolumn{3}{c}{0.829 (0.802, 0.868)}      & \multicolumn{3}{c}{0.188 (0.153, 0.243)}      & \multicolumn{3}{c}{0.049 (0.042, 0.055)}                               & \multicolumn{3}{c}{0.039 (0.033, 0.045)}                                         \\ \hline
\end{tabular}}
\end{table}

In summary, we have tested the performance on various simulation settings (model assumptions, event rates, sample sizes, non-linearity, {\it etc.}) where VIE takes the lead in all cases. IAF- and GPD-only variants perform poorly, even not comparable to the vanilla VAE model. This is possibly due to the prior is not matched. Explicitly matching the prior via the Fenchel mini-max scheme improves the performance, especially in the long-tailed representation datasets. Stacked together, our full proposal of VIE consistently outperforms its variants and always approaching the oracle performance in the large sample regime.
\subsection{D. Real-world datasets}\label{sup:real}
\vspace{3pt}
We consider 5 real-world datasets, including 3 survival datasets in the study. Among those dataset, \texttt{COVID} and \texttt{InP} are from Duke University Health System (DUHS), which are not public at this time. \texttt{SEER}\citepSM{ries2007seer} and \texttt{SLEEP}\citepSM{quan1997sleep} are two public survival datasets. Besides above clinical-based datasets, we further evaluate the model performance on  \texttt{Fraud} dataset \citepSM{dal2017credit} in this supplementary material.
\subsubsection{Baseline Models.} In all experiments, LDAM, FOCAL, IW, DeepSVDD and MLP are specified in terms of three-layer MLPs of 32 hidden units with ReLU activation. When tuning parameters for LASSO, based on the notation of \citetSM{scikit-learn} function \texttt{sklearn.linear\_model.Lasso}, we choose $\alpha$ from $[\num{e-5}, \num{e-4}, \num{e-3}, \num{e-2}, 0.1, 0.2,0.5,0.8]$ referred to the best performance on the validation datasets. In Focal Loss, the parameter $\gamma$ is selected from the list $[0.1, 0.5, 1.0, 1.5, 2.0]$ based on the best performance on the validation datasets. 
\subsubsection{COVID Dataset} The dataset includes inpatient encounters to DUHS as of January 1, 2020. Vitals, administered medications, lab results, comorbidities, etc. are used as predictors to identify the risk of inpatient death, ventilation, and ICU transfer as adverse outcomes.
The raw data's detailed description for each group of covariates can be found in Table~\ref{tab:covidcov}. The mortality rate in this dataset is $2.8\%$, ventilation $7.8\%$ and ICU transfer $18\%$. From rare event modeling purposes, apart from the mortality prediction, we set the group of patients who experienced either death or ventilation as the combined adverse outcome group, which has $8\%$ event rate.
\begin{table}[!htbp]
\centering
\caption{Raw \texttt{COVID} dataset covariates before pre-processing.}
\label{tab:covidcov}
\resizebox{.5\textwidth}{!}{\begin{tabular}{@{}lll@{}}
\toprule
Data Name                 & Data Type   & Number Covariates \\ \midrule
Demographics              & Numerical   & 1 (age)           \\
Previous Encounters       & Numerical   & 2                 \\
Prior Procedures w/n year & Categorical & 186               \\
Problem List w/n year     & Categorical & 273               \\
Comorbidities w/n year    & Categorical & 545               \\
Chief Complaint           & Categorical & 100               \\
Lab Analytes Collected    & Categorical & 44                \\
Lab Analytes Results      & Numerical   & 44                \\
Orders Placed             & Categorical & 32                \\
Medications Administered  & Categorical & 74                \\
Vitals Recorded           & Numerical   & 37                \\ \bottomrule
\end{tabular}}
\end{table}

\begin{figure}[h]
    \centering
    \includegraphics[width=0.8\linewidth]{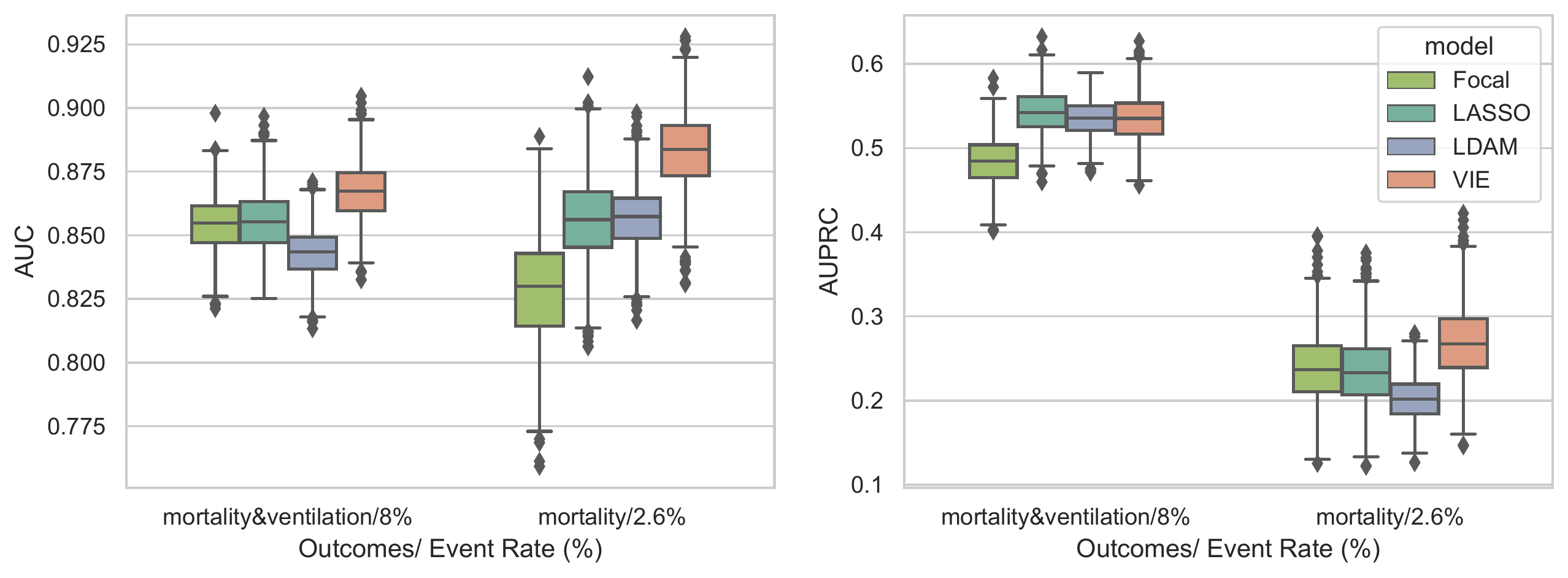}
\caption{Bootstrapped AUC (left) and AUPRC (right) Distribution of COVID datset with different outcomes. Note that comparisons of AUPRC among event-rates groups are meaningless.}
\label{fig:COVIDsummary}
\end{figure}

In Figure~\ref{fig:COVIDsummary}, we presented the comparison of VIE versus other baseline models with the two outcomes (combined and mortality). VIE shows strong performance under these metrics.

{\bf Cross-validation results}
To qualitatively show the superior performance of VIE, we examine the performances on a 5-fold cross validation of the \texttt{COVID} dataset for mortality prediction. VIE outperforms other baseline consistentlly on each fold. Comparing to the performance of the second-best model LDAM with a paired t test, the p-value yields $0.093 < 0.1$, with effect size 0.98, which shows the performance gap is statistically significant at $\alpha=0.1$.
\begin{table}[ht]
\centering
\caption{5-Fold cross validation results for COVID-19 dataset}
\label{tab:cvcovid}
\resizebox{0.8\textwidth}{!}{\begin{tabular}{@{}c|ccccc|ccccc@{}}
\toprule
       & \multicolumn{5}{c}{AUC}                                                            & \multicolumn{5}{c}{AUPRC}                                                          \\ \midrule
k-Fold & 1              & 2              & 3              & 4              & 5              & 1              & 2              & 3              & 4              & 5              \\ \midrule
LASSO  & 0.845          & 0.834          & 0.819          & 0.818          & 0.817          & 0.234          & 0.183          & 0.213          & 0.185          & 0.181          \\
VAE    & 0.831          & 0.842          & 0.848          & 0.800          & 0.764          & 0.198          & 0.181          & 0.191          & 0.196          & 0.147          \\
MLP    & 0.852          & 0.836          & 0.845          & 0.852          & 0.821          & 0.248          & 0.174          & 0.236 & 0.235          & 0.182          \\
Focal  & 0.847          & 0.837          & 0.836          & 0.851          & 0.836          & 0.229          & 0.158          & 0.216          & 0.241          & 0.186          \\
LDAM   & 0.842          & 0.848          & 0.843          & 0.839          & 0.839          & 0.240          & 0.197          & 0.234          & 0.209          & 0.182          \\
VIE    & \textbf{0.860} & \textbf{0.849} & \textbf{0.851} & \textbf{0.865} & \textbf{0.840} & \textbf{0.263} & \textbf{0.210} & \textbf{0.238}          & \textbf{0.256} & \textbf{0.201} \\ \bottomrule
\end{tabular}}
\end{table}

\subsubsection{InP Dataset} 
\begin{figure}[!htbp]
    \centering
    \includegraphics[width=.7\linewidth]{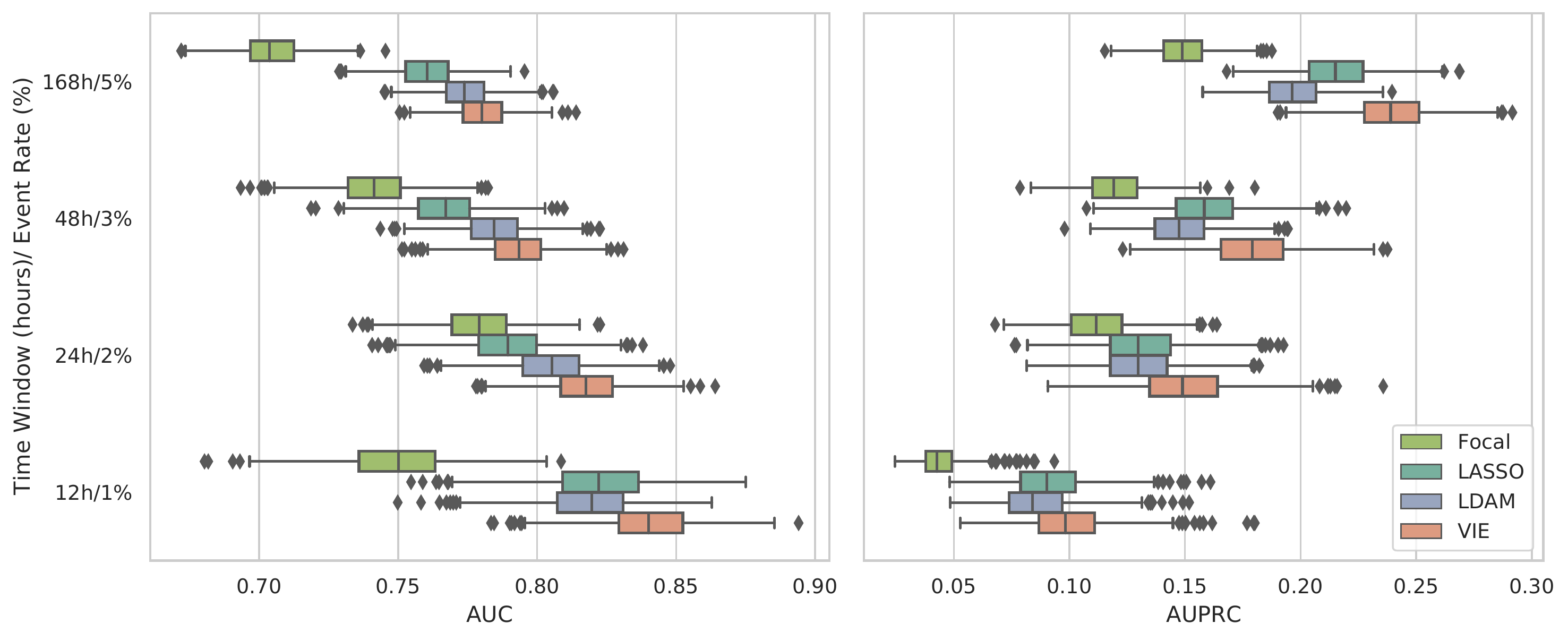}
\caption{Bootstrapped AUC (left) and AUPRC (right) Distribution of InP datset with different event rate. Note that comparisons of AUPRC among event-rates groups are meaningless.}
\label{fig:InPsummary}
\end{figure}

The dataset is another inpatient data of 82,450 Duke University Health System (DUHS) collected between 2014-2016. We abstracted time-varying clinical data ({\it i.e.}, vital signs, laboratory tests, medications) and followed patients until the Intensive Care Unit (ICU) transfer or
Death. We extracted their first encounter in the system (the admission) to generate this classification study to predict the risk of the occurrence of adverse outcomes (death or ICU transfer). The descriptions of raw data can be found in Table~\ref{tab:InPcov}. 
\begin{table}[!htbp]
\centering
\caption{Raw \texttt{InP} dataset covariates before pre-processing}
\label{tab:InPcov}
\resizebox{.5\textwidth}{!}{\begin{tabular}{@{}lll@{}}
\toprule
Data Name              & Data Type   & Number Covariates                \\ \midrule
Demographics           & Numerical   & 3 (age, sex, race)               \\
Admission Information  & Categorical & 2 (source and department)        \\
Vitals Recorded        & Numerical   & 10 (Diastolic, Resp, SpO2, etc.) \\
Lab Analytes Collected & Categorical & 30                               \\
Lab Analytes Results   & Numerical   & 30                               \\
Lab Orders Placed      & Categorical & 30               \\\hline
\end{tabular}}
\end{table}
With different sizes of the time windows, we generate four classification datasets with different event rates. As summarized in Figure~\ref{fig:InPsummary},
VIE takes a consistent lead in both AUC and AUPRC. The advantage enlarges when the event rates drop. Note that the trend of AUPRC when event rate shrinking is not meaningful\citep{boyd2012unachievable}.

\subsubsection{SEER and SLEEP Datasets} 
\texttt{SEER} and \texttt{SLEEP} are two public survival datasets that contain censoring (\textit{i.e.,} an event that is not reported during the follow-up period of a subject). To create a classification dataset from a survival dataset, we deleted patients censored before the time-cut. The proportion of subjects excluded for \texttt{SEER} is less than $0.1\%$, for \texttt{SLEEP} dataset is less than $0.2\%$, which should not affect the overall credibility of the analysis. We follow the pre-processing steps provided in \citetSM{chapfuwa2020survival}.
\begin{figure}[h]
\centering
\begin{minipage}{.4\textwidth}
  \centering
  \includegraphics[width=.86\linewidth]{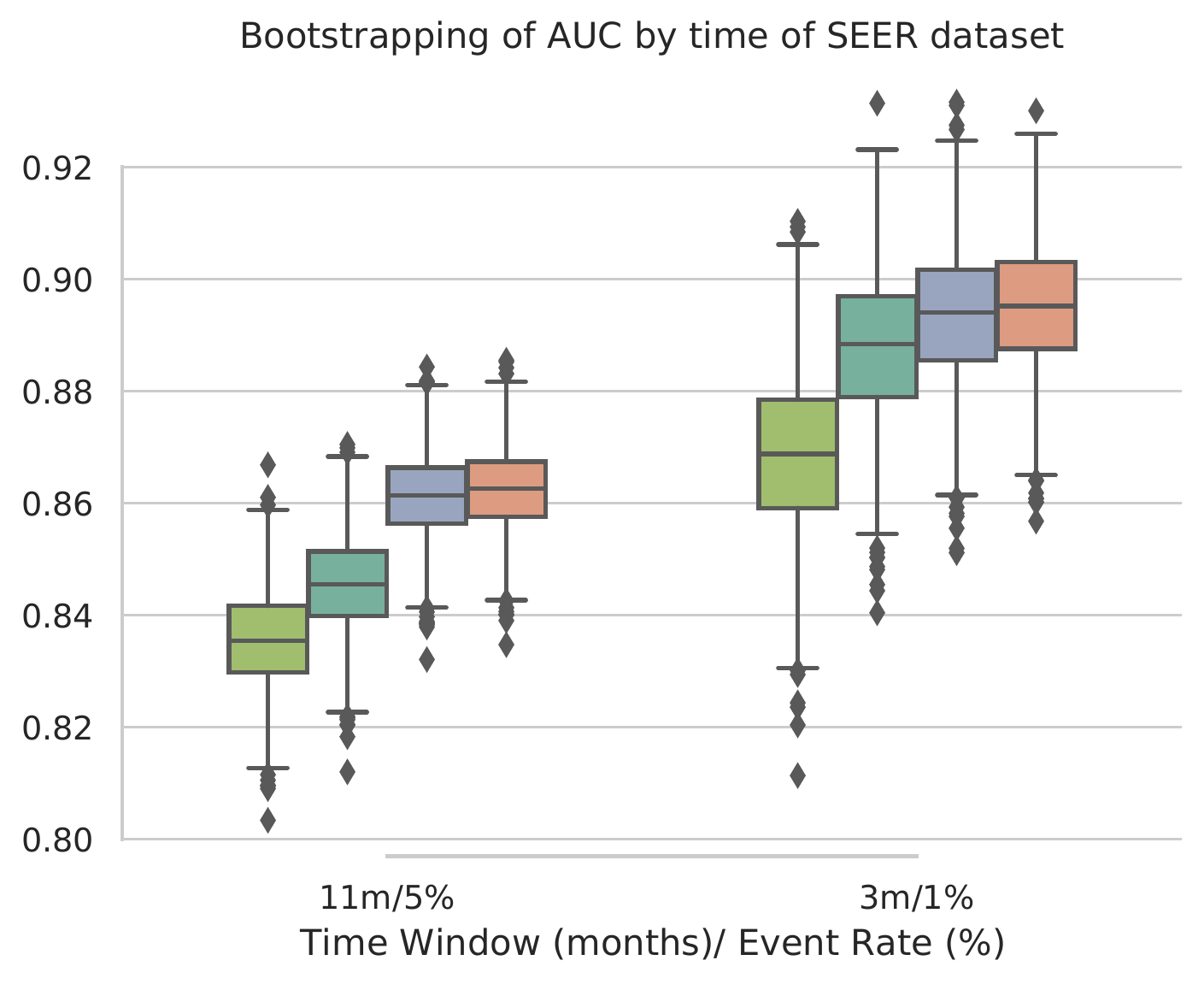}
\end{minipage}%
\begin{minipage}{.4\textwidth}
  \centering
  \includegraphics[width=.86\linewidth]{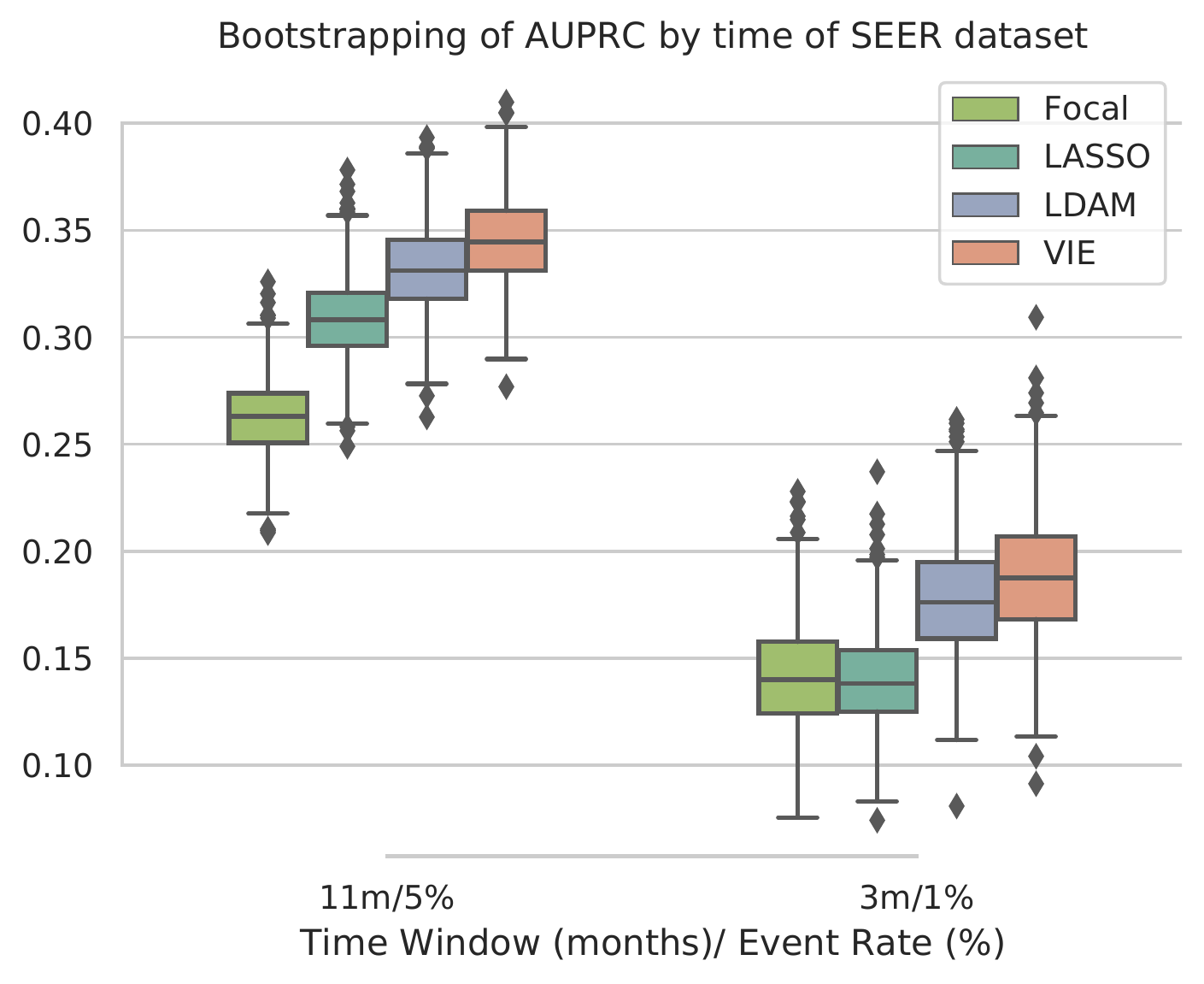}
\end{minipage}
\caption{Bootstrapped AUC (left) and AUPRC (right) Distribution of SEER dataset with different event rate.}
  \label{fig:SEERsummary}
\end{figure}
\begin{figure}[h]
    \centering
    \includegraphics[width=.72\linewidth]{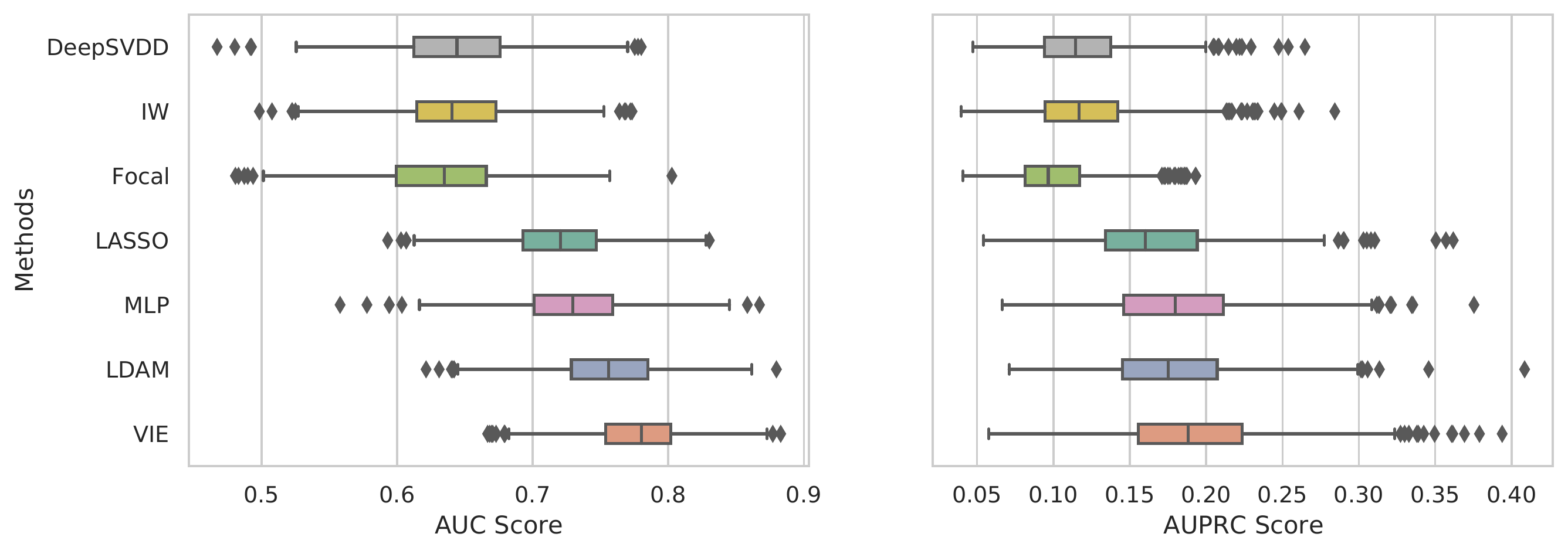}
\caption{Bootstrapped AUC (left) and AUPRC (right) Distribution of SLEEP dataset with $5\%$ event rate.}
\label{fig:SLEEPsummary}
\end{figure}
%


\subsubsection{Credit Card Fraud Detection} To evaluate the performances on non-clinical data, we examined the VIE model on fraud detection benchmark dataset \citepSM{dal2017credit}, where fraudulent credit card transactions are coined as rare events ($\sim 0.2\%$). The dataset includes 284k records with 29 covariates. We split the original dataset into training, validation, and testing datasets with a 6:2:2 ratio to ensure fair and stable comparison. The hyperparameters are selected based on the best performance on the validation dataset. The average and standard deviation of the metrics are presented in Table~\ref{tab:fraud}. VIE outperforms other baselines and achieved an average of over $0.99$ AUC in the bootstrapped samples.

\begin{table}[!htbp]
\centering
\caption{Fraud transaction classification.}
\label{tab:fraud}
\resizebox{.8\textwidth}{!}{
\begin{tabular}{@{}llllllll@{}}
\toprule
      & Lasso         & MLP           & DeepSVDD      & IW            & Focal          & LDAM          & VIE        \\ \midrule
AUC   & 0.981 (0.006) & 0.984 (0.007) & 0.796 (0.019) & 0.777 (0.026) & 0.916  (0.020) & 0.987 (0.005) & \textbf{0.991} (0.003) \\
AUPRC & 0.79 (0.032)  & 0.80 (0.030)  & 0.01 (0.002)  & 0.57 (0.039)  & 0.79 (0.032)   & 0.78 (0.037)  & \textbf{0.80} (0.032)  \\ \bottomrule
\end{tabular}}
\end{table}

\subsubsection{Exploration of the feature representation}
\begin{figure}[!htbp]
\centering
\begin{subfigure}{.43\textwidth}
  \centering
  \includegraphics[width=\linewidth]{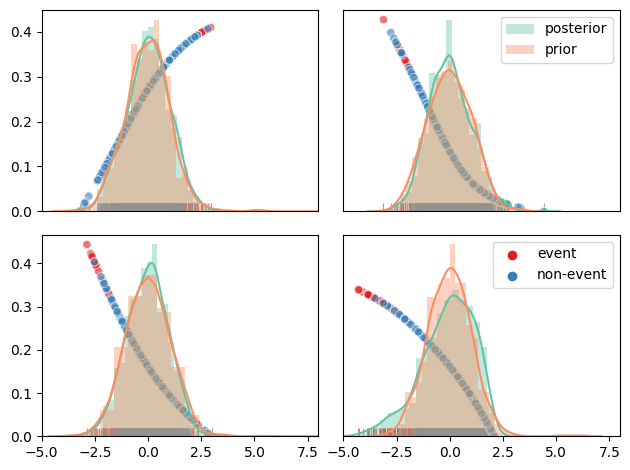}
  \caption{Learned prior and posterior distribution and monotonic predicted risks}
  \label{fig:InPer01hz}
\end{subfigure}%
\begin{subfigure}{.43\textwidth}
  \centering
  \includegraphics[width=\linewidth]{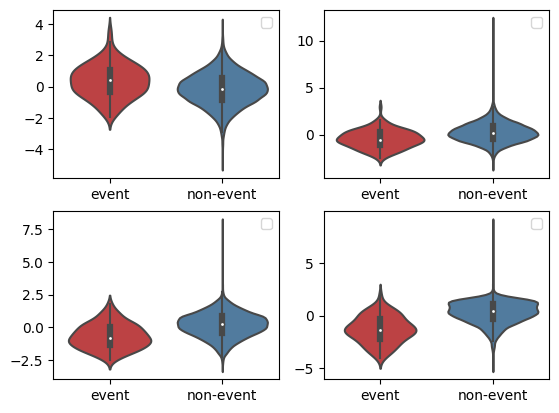}
  \caption{The latent representation values distribution grouped by event type}
  \label{fig:InPer01box}
\end{subfigure}
\caption{Four latent dimensions from the \texttt{InP} dataset (1\%) event rate, where the extreme distribution in the first dimension is the simulator to the events, the other three dimensions serve as inhibitors }
\label{fig:InPer01-latent}
\end{figure}

We visualize the marginal relationship between latent space dimensions and risk in the real-world dataset \texttt{InP} dataset ($1\%$ event rate), which are shown in Figure~\ref{fig:InPer01-latent}. The first dimension (top-left), the extremal behavior contributes significantly and positively to the event risk prediction. The other three dimensions serve as inhibitors to the event risk. Empirically, all the latent dimensions have a long-tailed distribution, with learned scale parameter $\xi > 0$.

\begin{figure}[h]
    \centering
    \includegraphics[width=.45\linewidth]{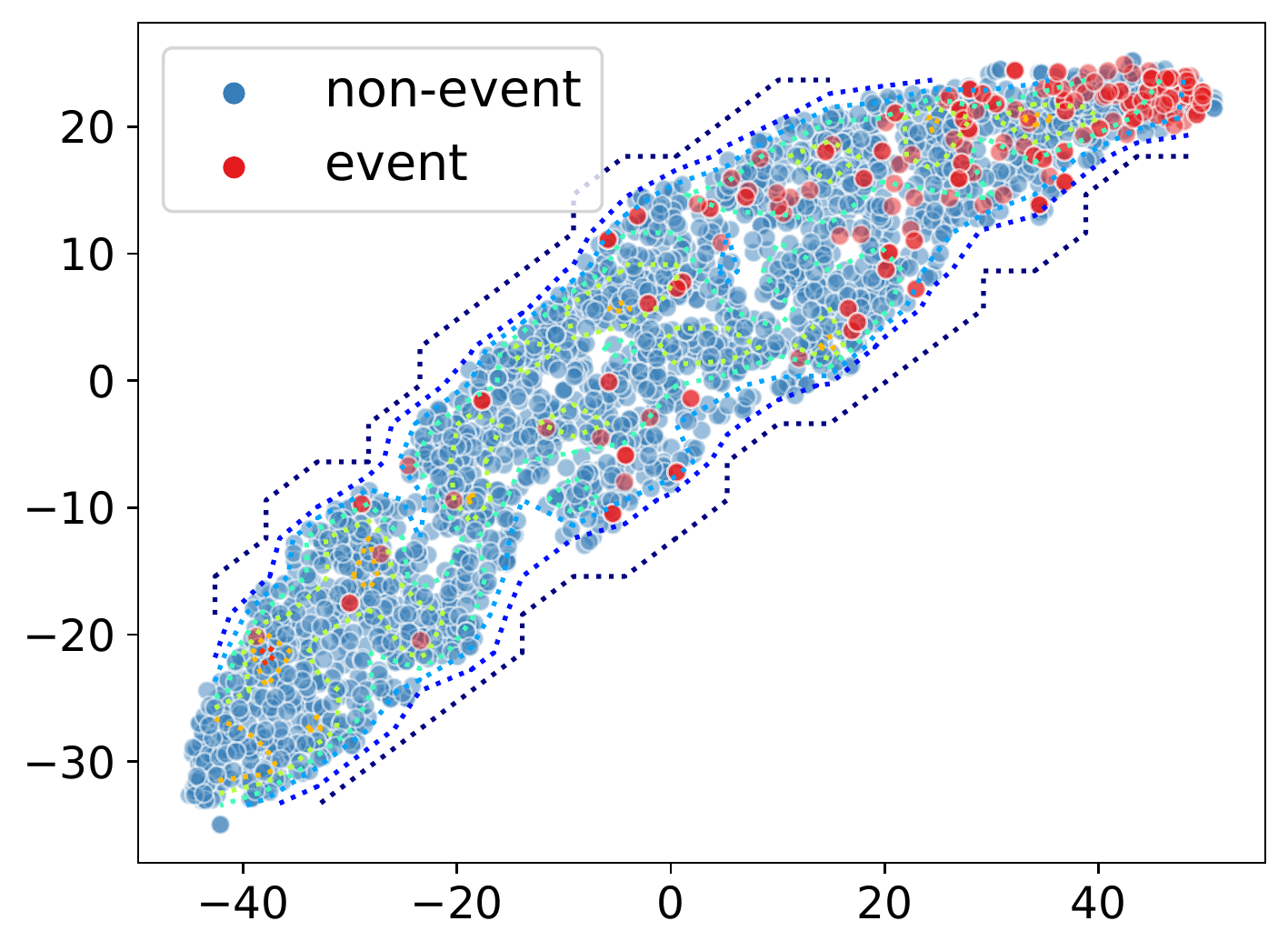}
    \caption{$t$-SNE plots with latent representation $z$.}
    \label{fig:InP-tSNE}
\end{figure}

We also embed the posterior space $z$ on a 2D plot with $t$-SNE, with probability contour lines, as shown in Figure~\ref{fig:InP-tSNE}. The events are concentrated to one end of the latent space.
\subsection{E. Generalized to Multiple-class classification}\label{sup:multiclass}
\begin{figure}[h]
    \centering
    \includegraphics[width=0.85\linewidth]{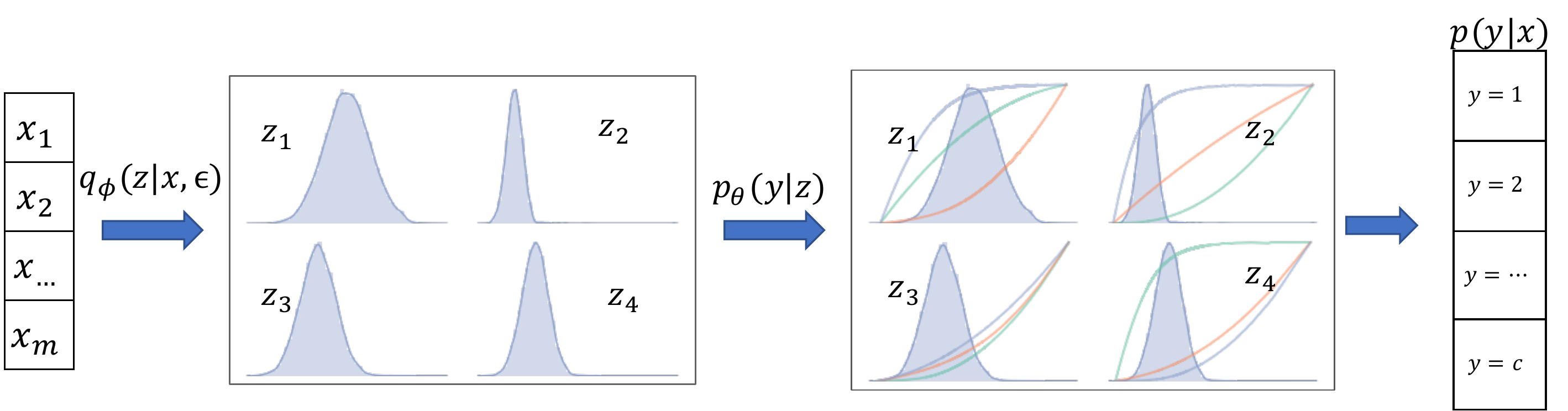}
    \caption{Illustration of multi-classification framework.}
    \label{fig:multiclass}
\end{figure}

Our binary classification framework can be generalized to multiple-class problems easily. 
We will stick to the mixed-GPD distribution of the posterior $z$ with $p$ dimensions, and increase the number of monotone networks for each dimension of $z$. In the binary case, each dimension of $z$ corresponds to one monotone network, here we can set it to $k$ networks per-dimension. In total, we now have $p\times k$ monotone functions. Then we can apply an FC layer to the final output, with $m$ categories, as shown in Figure~\ref{fig:multiclass}.
In the learning object, we would replace the binary cross-entropy loss (BCE) with regular cross-entropy loss (CE) in the reconstruction term.
\begin{align}
    \begin{aligned}
        p_{\theta}(y|z) & \leftarrow \Phi(H( z;\theta)) \text{ Soft-max } , \\
        H( z;\theta) & \leftarrow k\text{ Additive Monotone Neural Nets (\ref{eq:monotoneFunc-aprox})} \\
        p(z) & \leftarrow \text{Mixed GPD }(u, \xi_p, \sigma_p) \,\, (\ref{eq:mixedGPD}),  \\
        q_{\phi}(z|x) & \leftarrow \text{Inverse Autoregressive Flow } (\ref{eq:IAFposterior}), \\
        \nu(z) & \leftarrow \text{Standard neural network}.
    \end{aligned}
\end{align}

We generate a toy dataset to illustrate VIE's performance on multiclassification problems. Based on Algorithm~\ref{algo:simu1}, instead of setting a binary time-cut, now we split the generated time $t$ with a sequence of time-cuts based on the percentiles $[5\%, 15\%, 30\%, 60\%]$ of $t$. In this way, we have a dataset with $5$ categorical outcomes with event rates $[5\%, 10\%, 15\%, 30\%, 40\%]$, respectively. To evaluate the performance, except for the per-class accuracy, we use $F1$ score, which is a the harmonic mean of precision (True Positives) and recall (sensitivity), $\frac{2}{\text{recall}^{-1}+\text{precision}^{-1}}$, ranges from 0 to 1, where $1$ indicating better performance. We use micro-averaged F1-score (micro-F1) to calculate the overall F1 scores for all classes, 

\begin{table}[!htbp]
\centering
\caption{Performance on few-shots learning dataset}
\label{tab:fewshots}
\resizebox{.5\textwidth}{!}{\begin{tabular}{@{}lllllll@{}}
\toprule
            & class 1        & class 2        & class 3        & class 4        & class 5        & micro-F1        \\ \midrule
event rates & 5\%            & 10\%           & 15\%           & 30\%           & 40\%           &                 \\ \midrule
Focal       & \textbf{0.503} & 0.202          & 0.095          & 0.139          & 0.089          & 0.1368          \\
LDAM        & 0.012          & \textbf{0.321} & \textbf{0.228} & 0.165          & 0.594          & 0.3522          \\
VIE         & 0.054          & 0.054          & 0.046          & \textbf{0.372} & \textbf{0.823} & \textbf{0.4521} \\ \bottomrule
\end{tabular}}
\end{table}

Comparing to related methods: FOCAL and LDAM, the model VIE results on this toy example are comparable per class and better in terms of F1 score. FOCAL and LDAM are specified as 3-layer MLPs with 32 hidden units, and VIE uses the previous setting, except for $k=3$.


\bibliographystyleSM{apsr}
\bibliographySM{VIEVTref}